\documentclass[pdflatex,sn-mathphys-num,iicol]{sn-jnl}

\usepackage{graphicx}%
\usepackage{multirow}%
\usepackage{amsmath,amssymb,amsfonts}%
\usepackage{amsthm}%
\usepackage{mathrsfs}%
\usepackage[title]{appendix}%
\usepackage{xcolor}%
\usepackage{textcomp}%
\usepackage{manyfoot}%
\usepackage{booktabs}%
\usepackage{algorithm}%
\usepackage{algorithmicx}%
\usepackage{algpseudocode}%
\usepackage{listings}%
\usepackage{CJKutf8}%

\theoremstyle{thmstyleone}%
\theoremstyle{thmstyletwo}%

\theoremstyle{thmstylethree}%

\begin{document}
\begin{CJK}{UTF8}{min} 
\title[Article Title]{DKDS: A Benchmark Dataset of \underline{D}egraded \underline{K}uzushiji \underline{D}ocuments with \underline{S}eals for Detection and Binarization}


\author[1]{\fnm{Rui-Yang} \sur{Ju}}\email{jryjry1094791442@gmail.com}
\author[2]{\fnm{Kohei} \sur{Yamashita}}\email{yamashita.kohei.8j@kyoto-u.ac.jp}
\author[2]{\fnm{Hirotaka} \sur{Kameko}}\email{kameko@i.kyoto-u.ac.jp}
\author*[2]{\fnm{Shinsuke} \sur{Mori}}\email{forest@i.kyoto-u.ac.jp}

\affil[1]{\orgdiv{Graduate School of Informatics}, \orgname{Kyoto University}, \orgaddress{\street{Sakyo-ku}, \city{Kyoto}, \postcode{606-8501}, \country{Japan}}}
\affil[2]{\orgdiv{Academic Center for Computing and Media Studies}, \orgname{Kyoto University}, \orgaddress{\street{Sakyo-ku}, \city{Kyoto}, \postcode{606-8501}, \country{Japan}}}


\abstract{
Kuzushiji, a pre-modern Japanese cursive script, can currently be read and understood by only a few thousand trained experts in Japan.
With the rapid development of deep learning, researchers have begun applying Optical Character Recognition (OCR) techniques to transcribe Kuzushiji into modern Japanese.
Although existing OCR methods perform well on clean pre-modern Japanese documents written in Kuzushiji, they often fail to consider various types of noise, such as document degradation and seals, which significantly affect recognition accuracy.
To the best of our knowledge, no existing dataset specifically addresses these challenges. 
To address this gap, we introduce the Degraded Kuzushiji Documents with Seals (DKDS) dataset as a new benchmark for related tasks.
We describe the dataset construction process, which involves the assistance of a trained Kuzushiji expert, and define two benchmark tracks: (1) Kuzushiji character and seal detection and (2) document binarization.
For the Kuzushiji character and seal detection track, we provide baseline results using several recent versions of YOLO to detect Kuzushiji characters and seals.
For the document binarization track, we present baseline results from traditional binarization algorithms, traditional algorithms combined with K-means clustering, two state-of-the-art (SOTA) generative adversarial network (GAN) methods, and our improved conditional GAN (cGAN)-based method.
The DKDS dataset and the implementation code for baseline methods are available at \url{https://ruiyangju.github.io/DKDS}.}

\keywords{Kuzushiji, Seals, Pre-modern Japanese Documents, Benchmark Dataset, Kuzushiji Character Detection, Document Binarization}

\maketitle

\begin{figure*}[t]
\centering
\includegraphics[width=\linewidth]{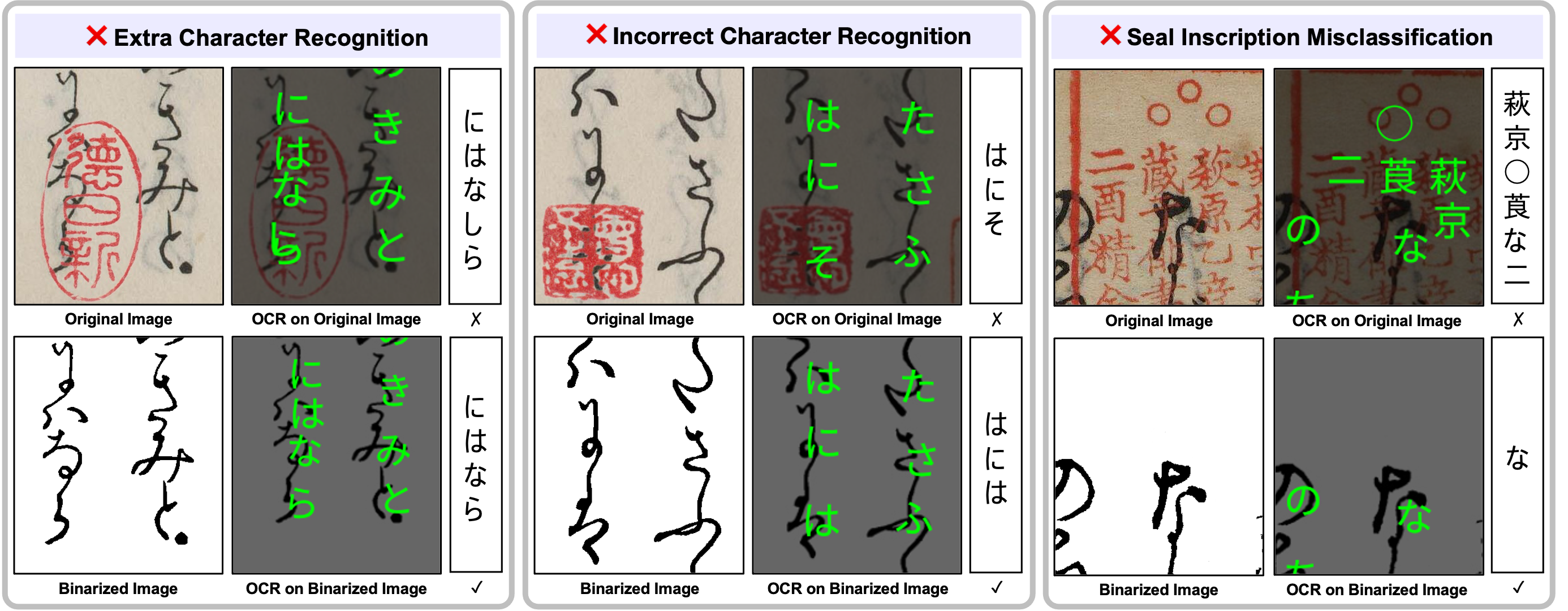}
\vspace{-8pt}
\caption{Comparison of Optical Character Recognition (OCR) results on Kuzushiji characters overlapping with seals between the original and binarized images. 
OCR was conducted using the ``miwo'' app~\cite{clanuwat2021miwo}.
From left to right, the observed OCR errors include recognition of extra character, recognition of incorrect character, and misclassification of seal inscriptions as text.}
\label{fig_motivation}
\vspace{-2pt}
\end{figure*}

\section{Introduction}
Pre-modern Japanese documents were mainly written in Kuzushiji\footnote{くずし字, a form of Japanese cursive script (草書体).}, a cursive script that was widely used in official records, personal letters, and literary works. 
Kuzushiji is characterized by complex, flowing strokes and highly variable forms, which differ markedly from modern Japanese characters~\cite{ueki2021survey}.
Educational reforms in the early 20th century introduced standardized textbooks and led to reduced instruction in Kuzushiji. 
Following the adoption of modern kana orthography in 1946, most contemporary native Japanese speakers are unable to read Kuzushiji characters without specialized training~\cite{hashimoto2017Kuzushiji}.

With remarkable advances in deep learning for computer vision, researchers have increasingly applied neural network models to Kuzushiji character recognition~\cite{toppan2023fuminoha,kiyonori2023enhancing,toru2024development}.
The primary datasets~\cite{clanuwat2018deep} for Kuzushiji character recognition include Kuzushiji-MNIST, Kuzushiji-49, and Kuzushiji-Kanji\footnote{Kanji refers to Chinese-origin characters used in Japanese.}. 
The Kaggle competitions~\cite{kitamoto2019progress,kitamoto2020kaggle} organized using these datasets have further advanced both research and practical applications~\cite{le2019human}.
Together, these datasets and competitions cover a wide range of Kuzushiji characters and provide a standardized evaluation platform for researchers.
With these resources, researchers can perform not only Kuzushiji character classification but also related tasks such as Kuzushiji character detection and generation~\cite{soan2023}, thereby promoting the development of Kuzushiji character recognition and improving model performance.

\begin{figure*}[t]
\centering
\includegraphics[width=\linewidth]{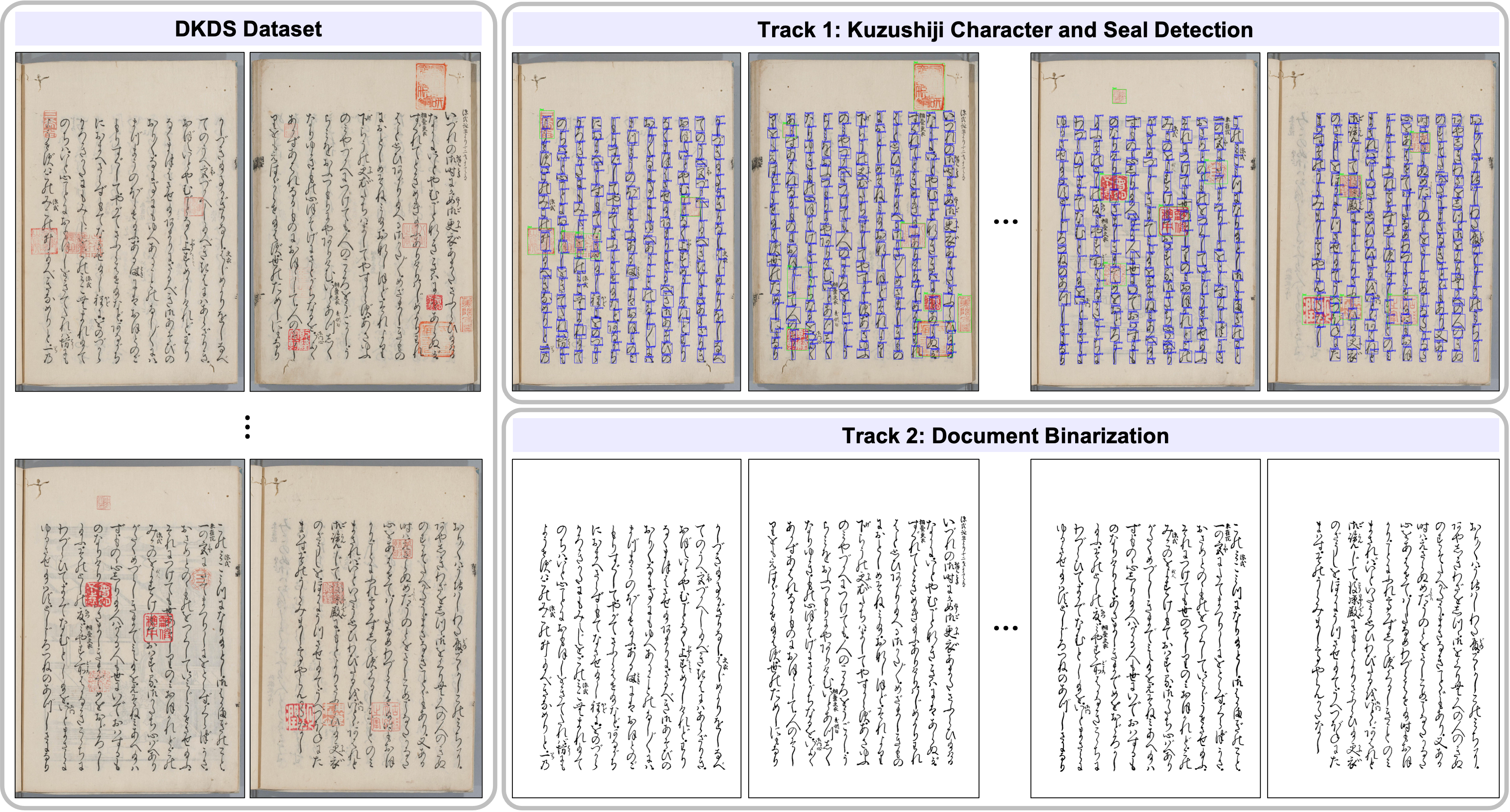}
\vspace{-8pt}
\caption{The proposed \textbf{DKDS} dataset is the first collection of degraded pre-modern Japanese document images specifically designed to address the challenge of Kuzushiji characters overlapping with seals.
Based on the dataset, we define two benchmark tracks: (1) Kuzushiji Character and Seal Detection, and (2) Document Binarization.}
\label{fig_intro}
\vspace{-2pt}
\end{figure*}

The Center for Open Data in the Humanities (CODH) has developed several Kuzushiji OCR systems that address both character-level and page-level recognition scenarios.
In 2021, CODH introduced the mobile application ``miwo''~\cite{clanuwat2021miwo}, which enables users to capture or upload images of Kuzushiji documents and automatically transcribe them into modern Japanese.
CODH also released KuroNet~\cite{clanuwat2019kuronet,lamb2020kuronet}, an end-to-end OCR model based on a Residual U-Net architecture~\cite{ronneberger2015u}.
Designed to capture long-range context, process extensive vocabularies, and accommodate non-standard character layouts, KuroNet supports robust recognition of pre-modern Japanese texts.
On Kaggle's Kuzushiji datasets~\cite{kitamoto2019progress,kitamoto2020kaggle}, KuroNet achieved classification accuracies exceeding 90\%.
More recently, SakanaAI introduced Metom~\cite{Metom}, a Vision Transformer~\cite{dosovitskiy2021vit}-based Kuzushiji classifier specialized for single-character recognition.
Metom was trained on 2,703 types of Kuzushiji characters that appeared at least five times in a large-scale Kuzushiji document dataset provided by CODH~\cite{genjimonogatari}, providing a powerful and extensible foundation for character-level analysis.

Existing OCR methods~\cite{toppan2023fuminoha,kiyonori2023enhancing,toru2024development,clanuwat2019kuronet,lamb2020kuronet} perform well on clean pre-modern Japanese documents written in Kuzushiji; however, they do not sufficiently consider the effects of noise, such as document degradation and seals, on recognition accuracy.
As shown in Fig.~\ref{fig_motivation}, such noise can cause errors that significantly reduce recognition accuracy.
To address this issue, document binarization is widely adopted as a preprocessing step because it preserves the foreground (black) while removing the background and noise (white), thereby effectively improving OCR performance.

Seals, serving as symbols of ownership and collector identity, are widely used in ancient and pre-modern Asian documents~\cite{li2021character}.
In these documents, seals usually appear as red marks inscribed with stylized ancient script and may contain the collector's name as well as phrases reflecting personal wishes, interests, or social status.
Among the various types of noise, the overlap between Kuzushiji characters and seals poses a major challenge for both Kuzushiji character and seal detection and document binarization.

To address this challenge, we introduce the Degraded Kuzushiji Documents with Seals (DKDS) dataset, developed in collaboration with a trained Kuzushiji expert.
This novel benchmark dataset comprises degraded pre-modern Japanese documents written in Kuzushiji, with various types of seals randomly added to simulate common forms of interference.
As shown in Fig.~\ref{fig_intro}, we define two task tracks for the DKDS dataset:
(1) Kuzushiji character and seal detection, and (2) document binarization.
The former track aims to accurately detect Kuzushiji characters and seal regions within the documents, while the latter focuses on preserving the primary textual content and removing noise, including stains and seals, providing clean input for subsequent OCR.

The contributions of this work are as follows:
\vspace{-6pt}
\begin{itemize}
\item[(a)] We introduce a novel benchmark dataset, named Degraded Kuzushiji Documents with Seals (DKDS), developed in collaboration with a trained Kuzushiji expert.
To the best of our knowledge, this is the first publicly available dataset specifically addressing the challenge of Kuzushiji characters overlapping with seals.
\item[(b)] We provide a detailed description of the dataset construction process, including the generation of binarization ground-truth, its verification and manual correction, and the random addition of seals to the raw Kuzushiji documents.
\item[(c)] We define the Kuzushiji character and seal detection task, which serves as a preliminary step for recognizing Kuzushiji characters and transcribing them into modern Japanese.
Baseline results for this task are also presented using several recent versions of YOLO models.
\item[(d)] We define the document binarization task specifically for pre-modern Japanese document images.
Compared to existing binarization datasets, the DKDS dataset presents additional challenges due to the overlap of Kuzushiji characters with seals.
Baseline results for this task are provided using traditional binarization algorithms, traditional algorithms combined with K-means clustering, two state-of-the-art (SOTA) GAN methods, and our improved conditional GAN (cGAN)-based method.
\end{itemize}

The rest of this paper is organized as follows.
Section~\ref{sec:related_works} reviews related datasets and previous work relevant to the two task tracks defined in this paper, highlighting their limitations.
Section~\ref{sec:dataset_construction} describes the dataset construction process, including raw data collection, generation of binarization ground-truth, and the procedure for adding seals.
Section~\ref{sec:track_1} introduces Track 1 (Kuzushiji Character and Seal Detection), including the task definition, evaluation metrics, and baseline methods.
Section~\ref{sec:experiment_1} presents the experimental setup and implementation details, followed by quantitative comparisons and visual results for Track 1.
Section~\ref{sec:track_2} introduces Track 2 (Document Binarization), including the task definition, evaluation metrics, and baseline methods.
Section~\ref{sec:experiment_2} presents the experimental setup and implementation details, followed by quantitative comparisons and visual results for Track 2.
Section~\ref{sec:ocr_evaluation} presents OCR evaluation results to further analyze the effectiveness of the proposed dataset and methods.
Finally, Section~\ref{sec:conclusion} concludes the paper and discusses future research directions.

\section{Related Work}\label{sec:related_works}
\subsection{Related Datasets}
The datasets relevant to this work can be categorized into three groups:
(1) Kuzushiji document datasets, (2) degraded document binarization datasets, and (3) seal datasets.

Existing research~\cite{clanuwat2018deep} on pre-modern Japanese documents written in Kuzushiji lacks a dataset that specifically addresses the challenge of overlap between characters and seals, which can significantly reduce recognition accuracy.

Most publicly available datasets for degraded document binarization focus on documents written in non-logographic writing systems.
For example, the documents in the Document Image Binarization Contest (DIBCO) series~\cite{gatos2009icdar,pratikakis2019icdar2019}, the Bickley Diary (BD)~\cite{deng2010binarizationshop}, and the Synchromedia Multispectral Ancient Document Images (SMADI)~\cite{hedjam2013historical} are primarily written in alphabetic scripts, whereas the Persian Heritage Image Binarization Dataset (PHIBD)~\cite{ayatollahi2013persian} contains documents written in an Abjad script. 
In contrast, to the best of our knowledge, there is currently no publicly available document binarization dataset specifically designed for pre-modern Japanese documents, particularly those written in Kuzushiji.

Furthermore, no existing degraded document datasets include seals as the primary source of interference.
Yang \emph{et al.}~\cite{yang2023docdiff} synthesized paired data by overlaying seal images onto clean documents; however, their synthesized documents primarily contain modern Chinese characters, which differ significantly from the pre-modern Japanese Kuzushiji characters targeted in this work.
Following a similar strategy, we also simulate seal interference by randomly adding seal images to collected pre-modern Japanese documents.

Regarding the sources of seal images, existing publicly available seal datasets are unsuitable for this purpose, as they either do not contain ancient Asian seals, suffer from poor image quality, or exhibit complex backgrounds.
For instance, MiikeMineStamps~\cite{buitrago2021miikeminestamps} includes some Japanese seal images; however, they are generally low-resolution and feature non-uniform backgrounds, making background removal difficult.

\subsection{Kuzushiji Character and Seal Detection}
The diverse shapes of ancient seals, their curved inscriptions, and the frequent overlap between Kuzushiji characters and seals pose significant challenges for both seal detection and Kuzushiji character detection and recognition.
Accurate Kuzushiji character detection is a critical preliminary step for Kuzushiji character recognition.

For instance, DBNet~\cite{liao2020real}, a widely used text detection method in OCR systems, achieved an F-measure of 85.4\% at 26 frames per second on the ICDAR 2015 benchmark~\cite{karatzas2015icdar}, indicating both high detection accuracy and real-time performance.

Similarly, seal detection is important for subsequent seal analysis.
However, the scarcity of ancient seal samples and the prevalence of low-quality or degraded images severely limit the available training data and annotations for related tasks~\cite{li2021character}.
Micenkov \emph{et al.}~\cite{micenkov2011stamp} proposed an automatic seal segmentation system for insurance documents.
When evaluated on the collected StaVer dataset, which contains 400 document images, their system achieved a recall of 83\% and a precision of 84\%.
Yu \emph{et al.}~\cite{yu2023icdar} developed a modern seal dataset for official and financial contexts, introducing two tasks: seal title text detection and end-to-end seal title recognition. 

To the best of our knowledge, no dataset currently exists for Kuzushiji character and seal detection in degraded pre-modern Japanese documents written in Kuzushiji.

The YOLO series of models~\cite{jocher2023yolov8,wang2024yolov9,wang2024yolov10,jocher2024yolo11} achieve an excellent balance between detection accuracy and model size, demonstrating outstanding performance across a wide range of object detection tasks.
Researchers have explored the use of the YOLO series of models for Kuzushiji character detection and seal detection.
Nevertheless, due to the lack of ancient seal data from pre-modern Japanese documents, recent YOLO models cannot be directly applied to this specific scenario.

\subsection{Document Binarization}
Document binarization is the task of classifying each pixel in a document image as either foreground (text) or background (including noise), with foreground pixels represented in black and background pixels in white.
It is an important preprocessing step for OCR, as it effectively removes various degradations and visual artifacts~\cite{deng2010binarizationshop,hedjam2013historical,ayatollahi2013persian}, such as paper yellowing, text fading, ink bleeding, stains, and overlapping seals.

Early work in this area primarily relied on threshold-based methods.
Specifically, Otsu~\cite{otsu1979threshold} introduced a global thresholding algorithm in 1979.
Subsequently, Niblack~\cite{niblack1985introduction} and Sauvola~\cite{sauvola2000adaptive} proposed local adaptive thresholding techniques in 1985 and 2000, respectively, enabling more robust binarization of documents with non-uniform illumination and complex background variations.

To facilitate systematic evaluation, the Document Image Binarization Contest (DIBCO), one of the most popular competitions in this field, released benchmark datasets from 2009 to 2019~\cite{gatos2009icdar,pratikakis2019icdar2019}.
These datasets have played an important role in advancing both traditional and deep learning-based binarization methods by providing standardized evaluation benchmarks.

The introduction of GANs~\cite{goodfellow2020generative} has enabled the generation of high-quality binary document images.
Suh \emph{et al.}~\cite{suh2022two} proposed a two-stage GAN framework that employed six enhanced CycleGANs~\cite{zhu2017unpaired} to produce binarized document images.
Based on this work, Ju \emph{et al.}~\cite{ju2023ccdwt,ju2024three} further proposed a three-stage GAN architecture composed of six improved CycleGAN modules to further improve document binarization performance.
Both methods consistently achieved SOTA performance on the DIBCO benchmarks.

However, neither traditional binarization methods nor SOTA GAN-based methods have been evaluated on degraded pre-modern Japanese documents written in Kuzushiji, particularly those containing seals, overlaps between characters and seals, and complex visual degradations.
This gap highlights the need to develop datasets and models that are tailored to the unique characteristics of pre-modern Japanese documents.

\begin{figure}[t]
\centering
\includegraphics[width=\linewidth]{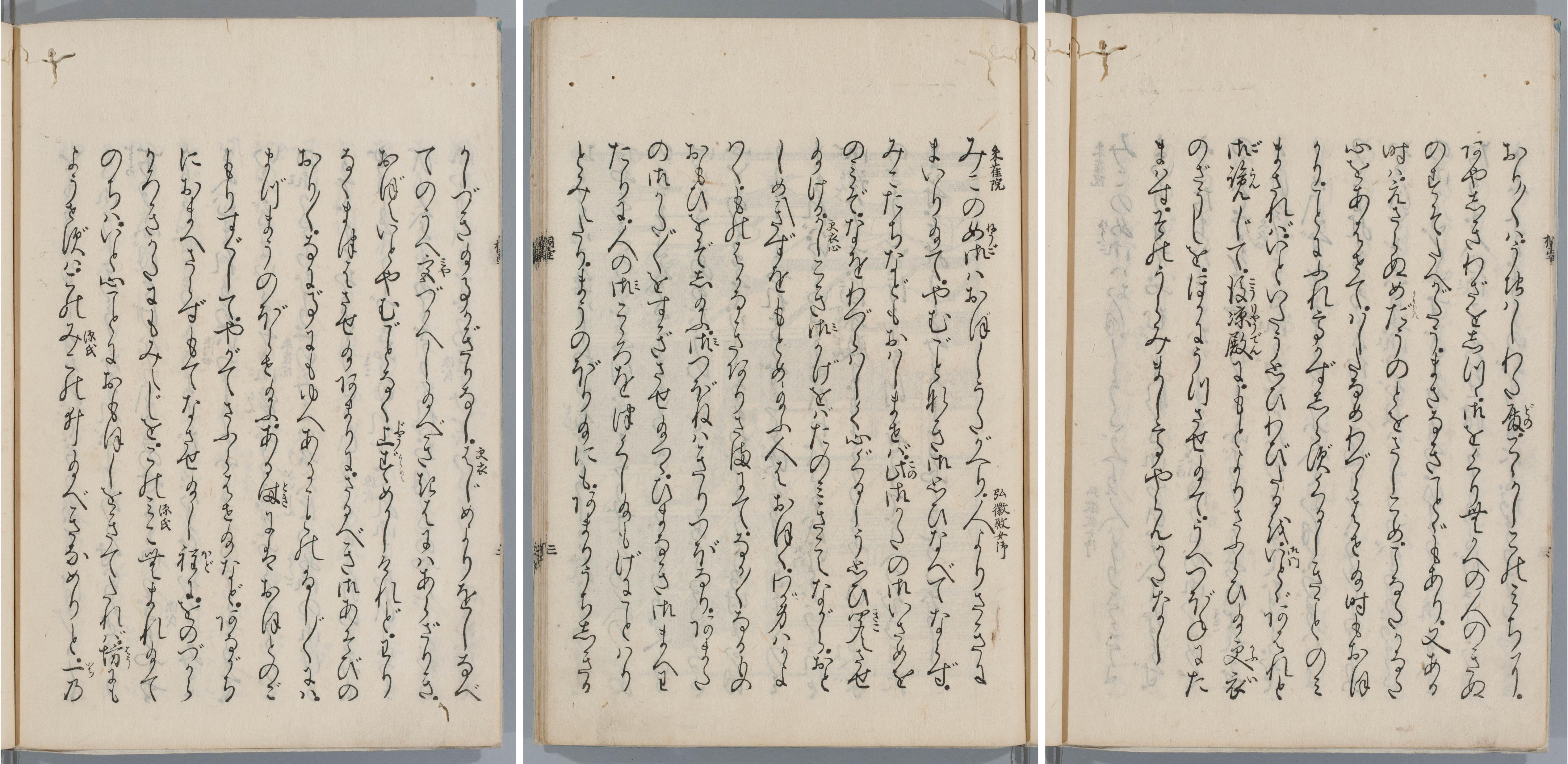}
\vspace{-8pt}
\caption{Examples of raw Kuzushiji document images from the book \emph{Genji Monogatari} (\emph{The Tale of Genji})~\cite{genjimonogatari}.}
\label{fig_rawdoc}
\vspace{-2pt}
\end{figure}

\section{Dataset Construction}\label{sec:dataset_construction}
\subsection{Raw Data Collection}
The raw Kuzushiji document images were obtained from the CODH~\cite{genjimonogatari}. 
Specifically, we selected the book \emph{Genji Monogatari} (\emph{The Tale of Genji}), with ID ``200003803''\footnote{https://codh.rois.ac.jp/char-shape/book/200003803/}, to construct the training set and the easy-level (Testing-E) and difficult-level (Testing-D) subsets of the DKDS test set.
These Kuzushiji document images have a typical resolution of approximately $2100\times3200$ pixels.
Examples of raw document images from this book are shown in Fig.~\ref{fig_rawdoc}.
This dataset was chosen for two main reasons:
(1) its historical significance and widespread popularity; and
(2) the availability of detailed OCR annotations provided by CODH, including 237 unique character types and a total of 11,132 characters.

\begin{figure}[t]
\centering
\includegraphics[width=\linewidth]{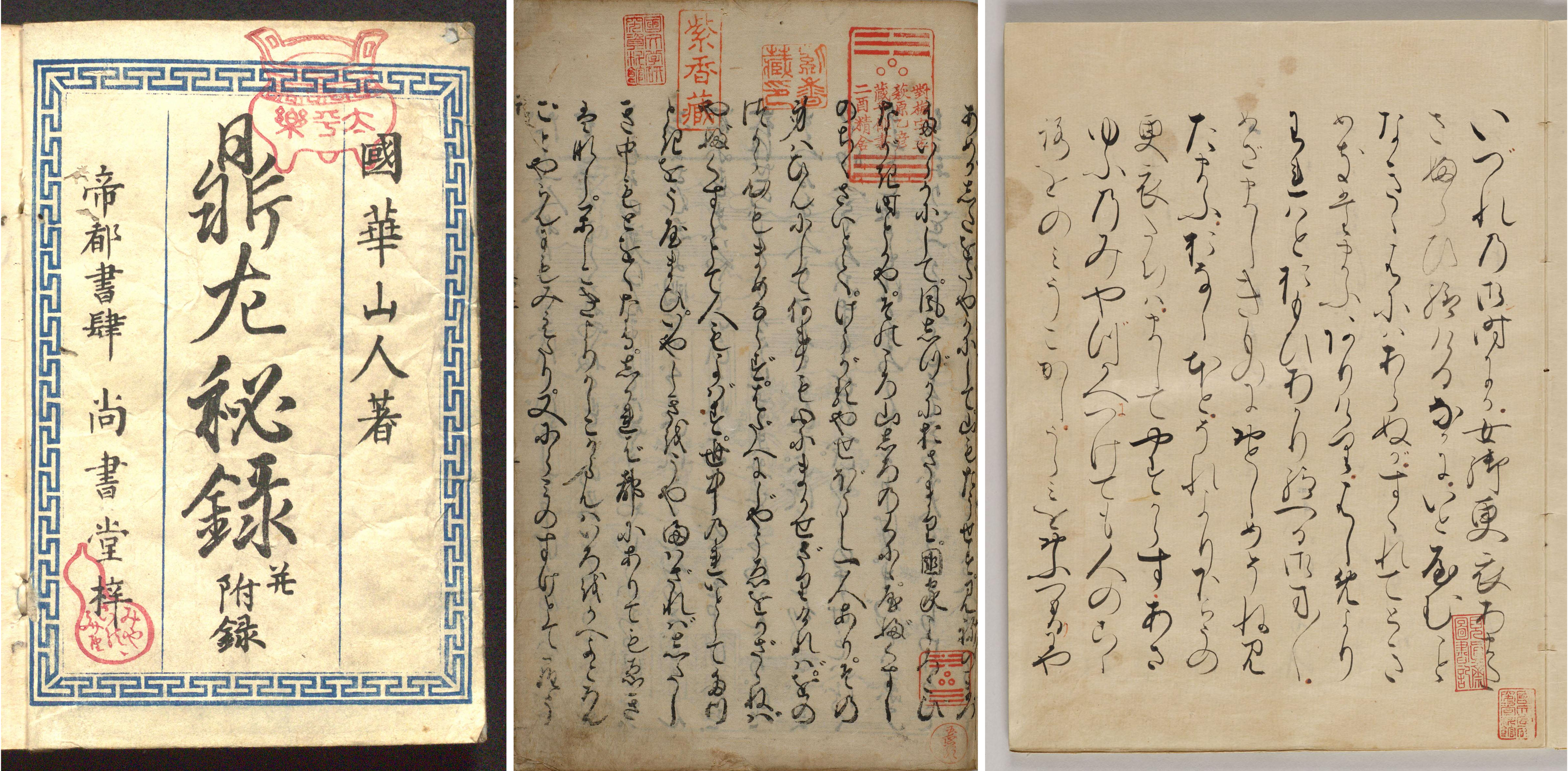}
\vspace{-8pt}
\caption{Examples of real Kuzushiji document images from the CODH~\cite{genjimonogatari} used to construct the Testing-R subset.}
\label{fig_real}
\vspace{-2pt}
\end{figure}

In addition, due to the limited number of documents with seals in the \emph{Genji Monogatari} dataset, we further selected 12 document images with seals from the CODH Kuzushiji dataset to construct the real-document (Testing-R) subset of the DKDS test set.
Examples from Testing-R are shown in Fig.~\ref{fig_real}.
The Testing-R is primarily used to evaluate trained models in real-world scenarios and to assess their robustness across different layouts, Kuzushiji writing styles, and seal shapes.

Furthermore, to date, there is no publicly available high-resolution dataset of ancient Japanese seals. 
To address this gap, we collected 219 seal images associated with three ancient Chinese emperors of the Qing dynasty (i.e., Kangxi (1661--1722), Yongzheng (1722--1735), and Qianlong (1735--1796)), provided by the developer of an ancient seal application~\cite{chen2023ancient}.
The primary reason for selecting ancient Chinese seals is that both ancient Japanese and Chinese seals are mainly inscribed with Kanji characters, and high-resolution images of ancient Japanese seals are currently unavailable.
Considering factors such as image resolution and seal patterns (e.g., whether the inscriptions are written in Kanji), we selected a final set of 128 seal images from the collection. 

\begin{figure}[t]
\centering
\includegraphics[width=\linewidth]{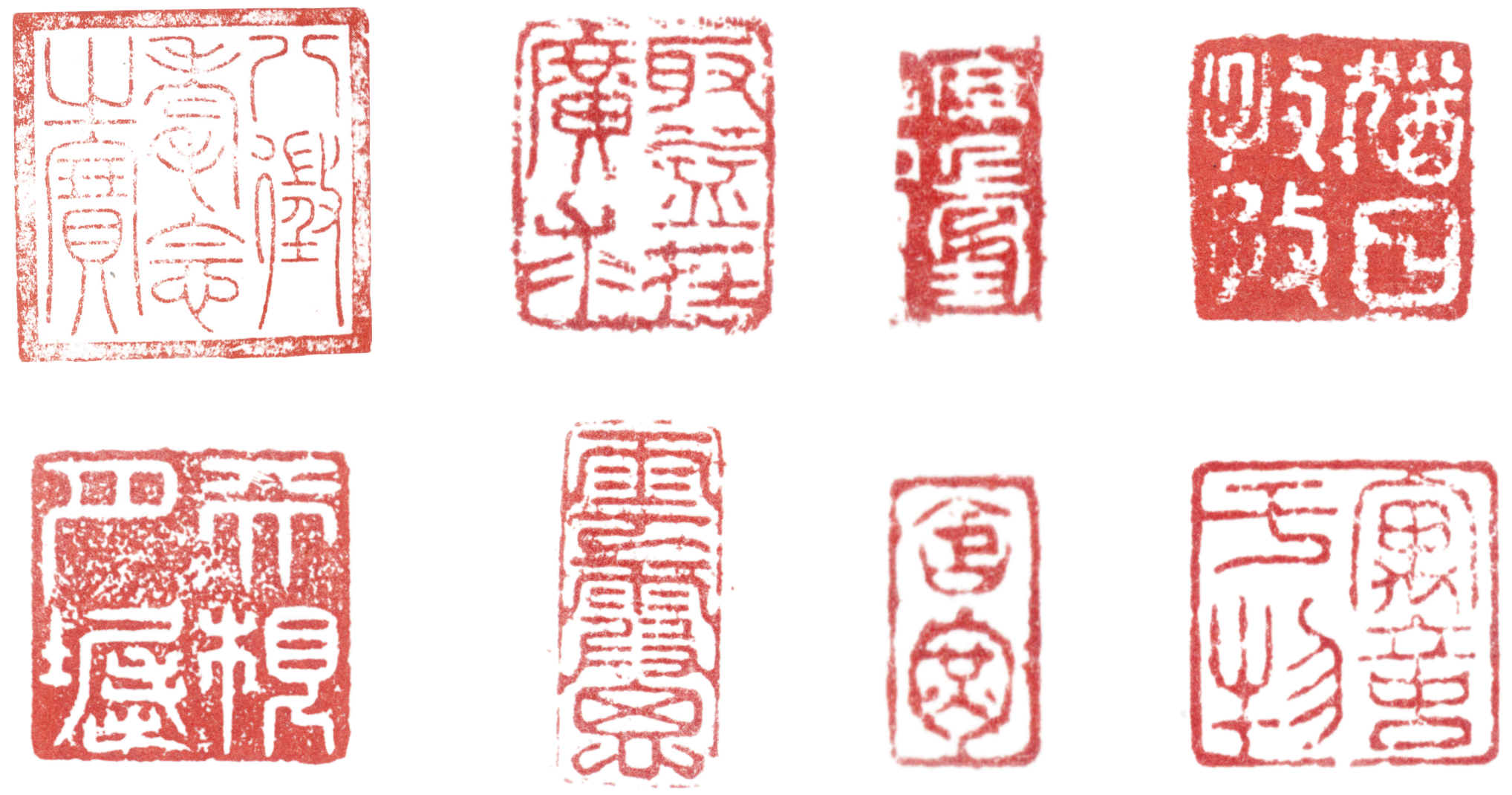}
\vspace{-8pt}
\caption{Examples of the collected imperial seals from the Qing dynasty, with backgrounds removed, which are used to simulate seal interference.}
\label{fig_rawseal}
\vspace{-2pt}
\end{figure}

As shown in Fig.~\ref{fig_rawseal}, these seal images are high-resolution and have easily removable backgrounds.
The background-removed seal images were saved in PNG format with transparent backgrounds and were loaded in RGBA format, where the alpha channel directly provided the transparency information used for image synthesis. 
No additional manual alpha estimation was performed.

\begin{figure*}[t]
\centering
\includegraphics[width=\linewidth]{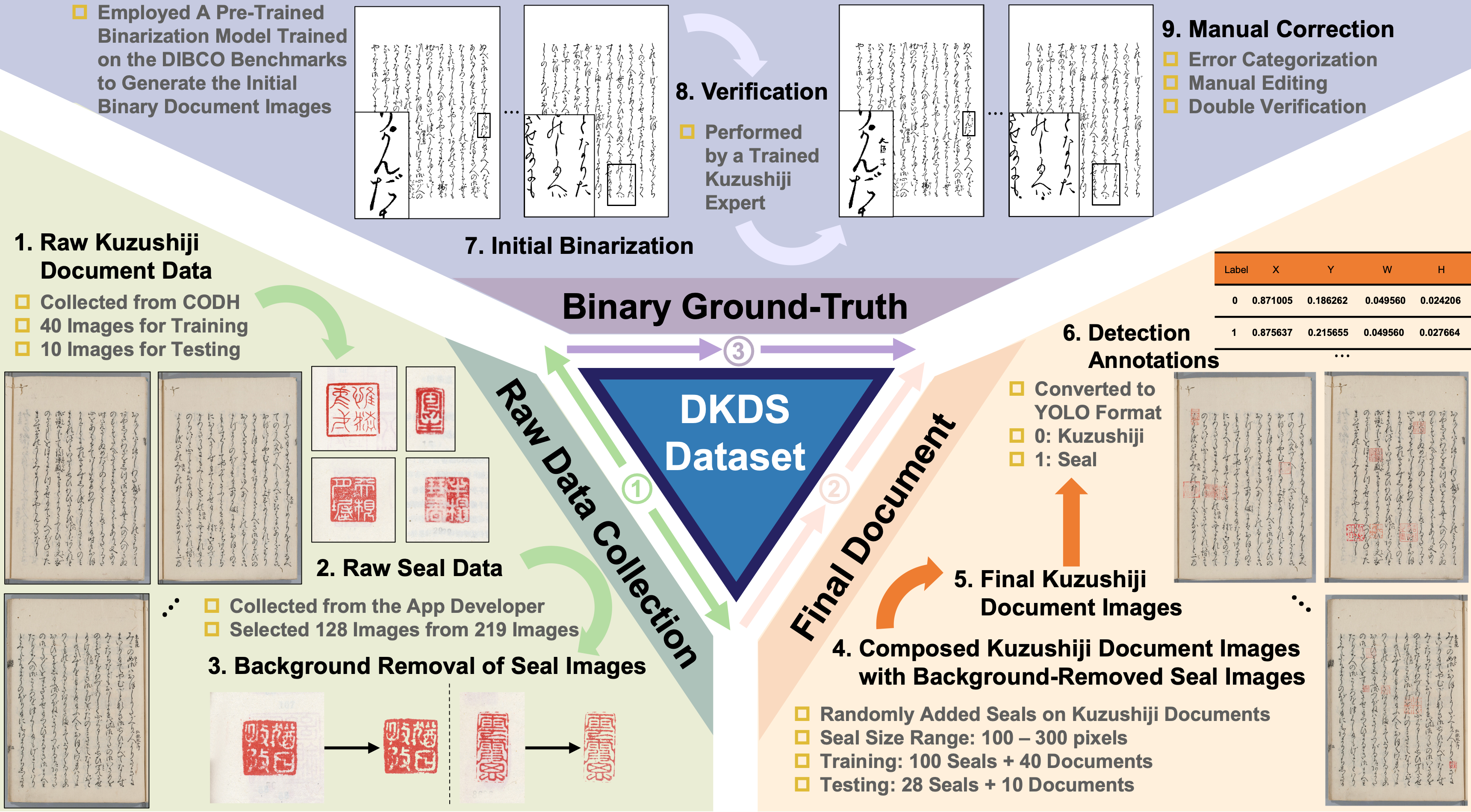}
\vspace{-8pt}
\caption{The overall workflow of the proposed \textbf{DKDS} dataset construction includes raw data collection, annotation for Kuzushiji character and seal detection, initial binarization ground-truth generation, verification, and manual correction.
The initial binarization ground-truth was generated using a pre-trained binarization model trained on the DIBCO benchmarks, while the verification and re-verification were conducted by two researchers, including a trained Kuzushiji expert.}
\label{fig_workflow}
\vspace{-2pt}
\end{figure*}

\begin{figure}[t]
\centering
\includegraphics[width=\linewidth]{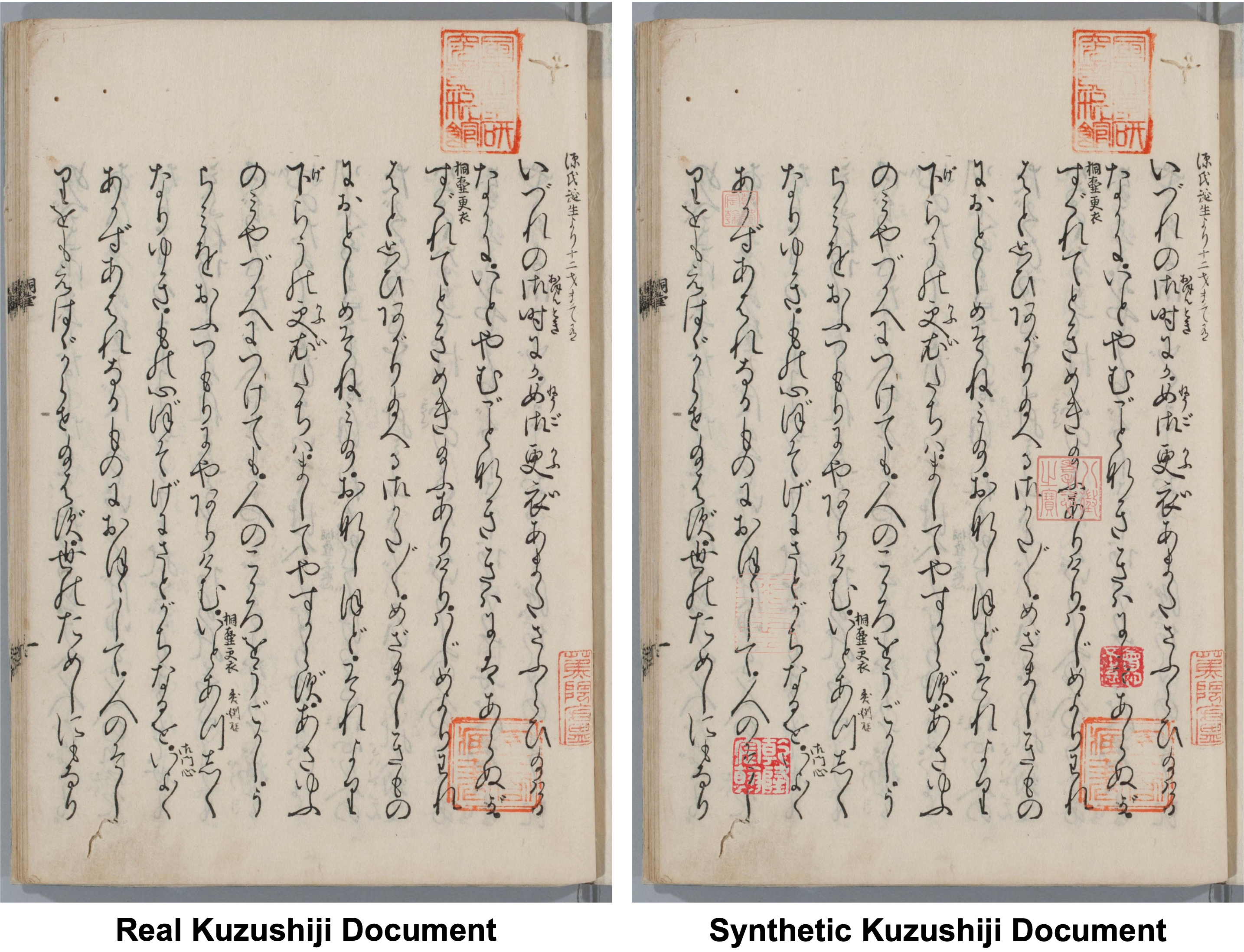}
\vspace{-8pt}
\caption{Visual comparison between seals in real Kuzushiji documents and those in the corresponding synthetic ones.}
\label{fig_synthetic}
\vspace{-2pt}
\end{figure}

\subsection{Construction Workflow}
For the training set and the Testing-E and Testing-D subsets, we constructed the dataset by producing synthetic color document images and their corresponding binarized ground-truth images from the raw data.
The overall workflow is shown in Fig.~\ref{fig_workflow}.
To create the final document images, we combined 128 background-removed seal images with 50 Kuzushiji document images.

For each document image, one or more seal images were selected and overlaid onto the document. 
Each seal image was first cropped to its non-empty bounding box, then randomly resized within a predefined scale range, and finally placed at a random position within a specified region of the document page. 
The bounding box of each inserted seal was recorded as an annotation during this process.
Specifically, we employed RGBA alpha compositing to overlay the seal images onto the Kuzushiji document images, such that the seal pixels were blended with the underlying document pixels according to the alpha channel values of the seal images.
This avoided hard pixel replacement and more realistically simulated the visual appearance of real seals partially covering handwritten characters.

In total, 40 Kuzushiji documents were selected for training, and the remaining 10 documents were used for testing.
As detailed in Table~\ref{tab:statistics}, the 128 seal images were divided into 100 for the training set (40 Kuzushiji documents) and 28 for the test set (10 Kuzushiji documents).

To increase the number of overlaps between seals and Kuzushiji characters in the training set, we duplicated the 100 training seal images to create a total of 200 seals, which were then randomly added to the 40 training Kuzushiji documents.
Specifically, each seal image was randomly resized within a predefined range (i.e., 100--300 pixels).
The seals were placed within the main content area of the document, defined as a rectangular region anchored at the top-left corner (x=100 pixels, y=120 pixels) with a width of 1900 pixels and a height of 2900 pixels, which approximately corresponds to the central area of the image.

Furthermore, we introduced an overlap control strategy to prevent excessive overlap between seals. 
Specifically, for each newly placed seal, we computed the overlap between its bounding box (x, y, width, height) and those of previously placed seals; if it overlapped with more than two existing seals, its position was re-sampled. 
This strategy prevents unrealistic stacking of multiple seals in highly congested regions.
As a result, each Kuzushiji document image in the training set contains an average of 5.0 seals.

\begin{figure*}[t]
\centering
\includegraphics[width=\linewidth]{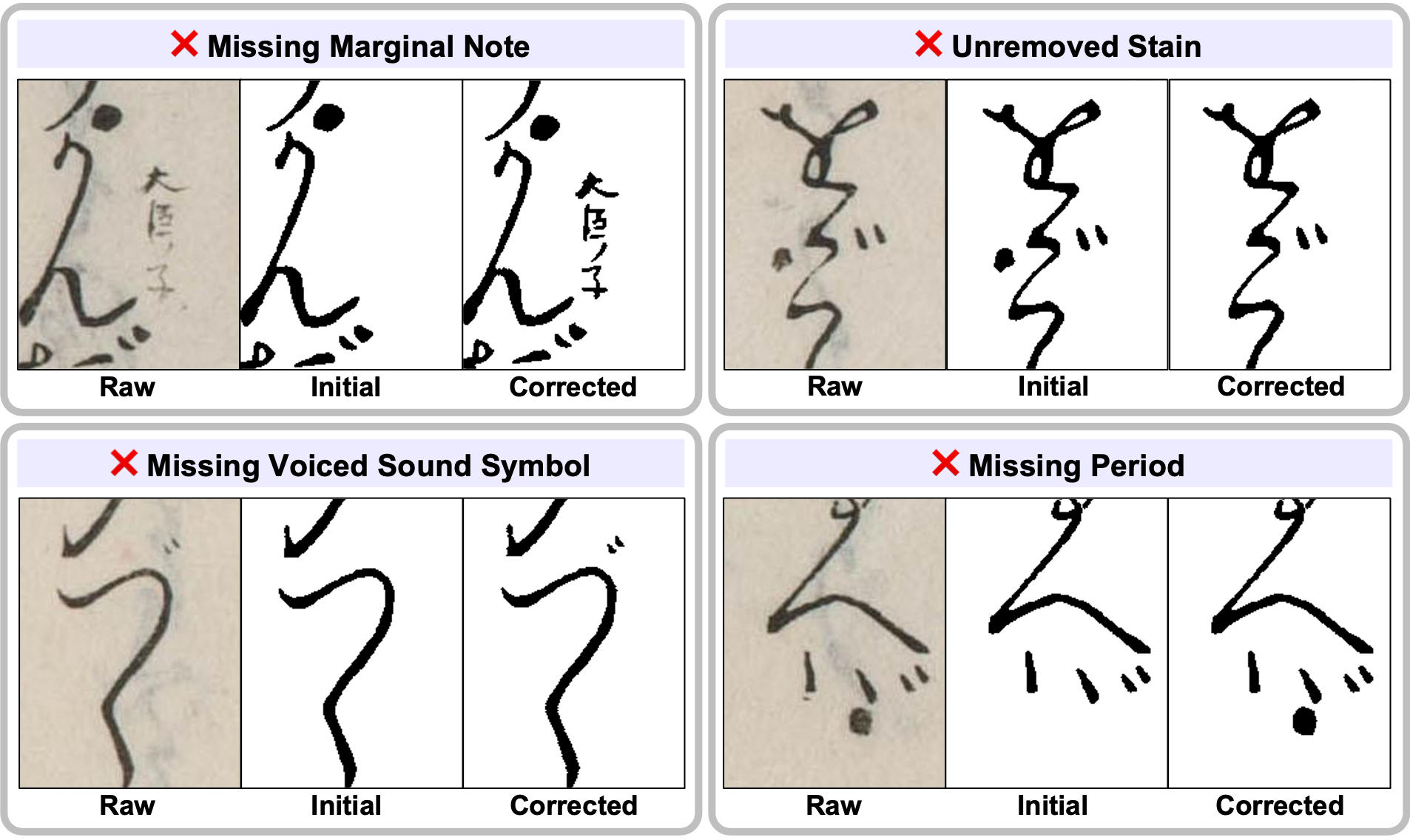}
\vspace{-8pt}
\caption{Examples of verification and manual correction during dataset construction.
``Raw'' indicates the input images, ``Initial'' denotes the binarized images produced by a pre-trained binarization model, and ``Corrected'' represents the final ground-truth images after verification and manual correction. 
Verification and re-verification were performed by two researchers, including a trained Kuzushiji expert, to determine which elements should be preserved (e.g., marginal notes, voiced sound symbols, and periods) and which should be removed (e.g., stains).}
\label{fig_verification}
\vspace{-2pt}
\end{figure*}

\begin{table*}[t]
\centering
\caption{Statistics of the proposed DKDS dataset, including the training set and test subsets. 
Testing-E, Testing-D, and Testing-R denote the ``Easy'', ``Difficult'', and ``Real'' subsets of the test set, respectively.}
\setlength{\tabcolsep}{9pt}{
\begin{tabular}{lccccc}
\toprule
\textbf{Subset} & \textbf{Documents} & \textbf{Characters} & \textbf{Characters/Doc} & \textbf{Seals} & \textbf{Seals/Doc} \\ \midrule
Training & 40 & 9,139 & 228.5 & 200 (100$\times$2) & 5.0 \\
Testing-E & 10 & 2,145 & 214.5 & 28 & 2.8 \\
Testing-D & 10 & 2,145 & 214.5 & 140 (28$\times$5) & 14.0 \\
Testing-R & 12 & 2,181 & 181.8 & 22 & 1.8 \\ \bottomrule
\end{tabular}}
\label{tab:statistics}
\vspace{-2pt}
\end{table*}

\begin{figure}[t]
\centering
\includegraphics[width=\linewidth]{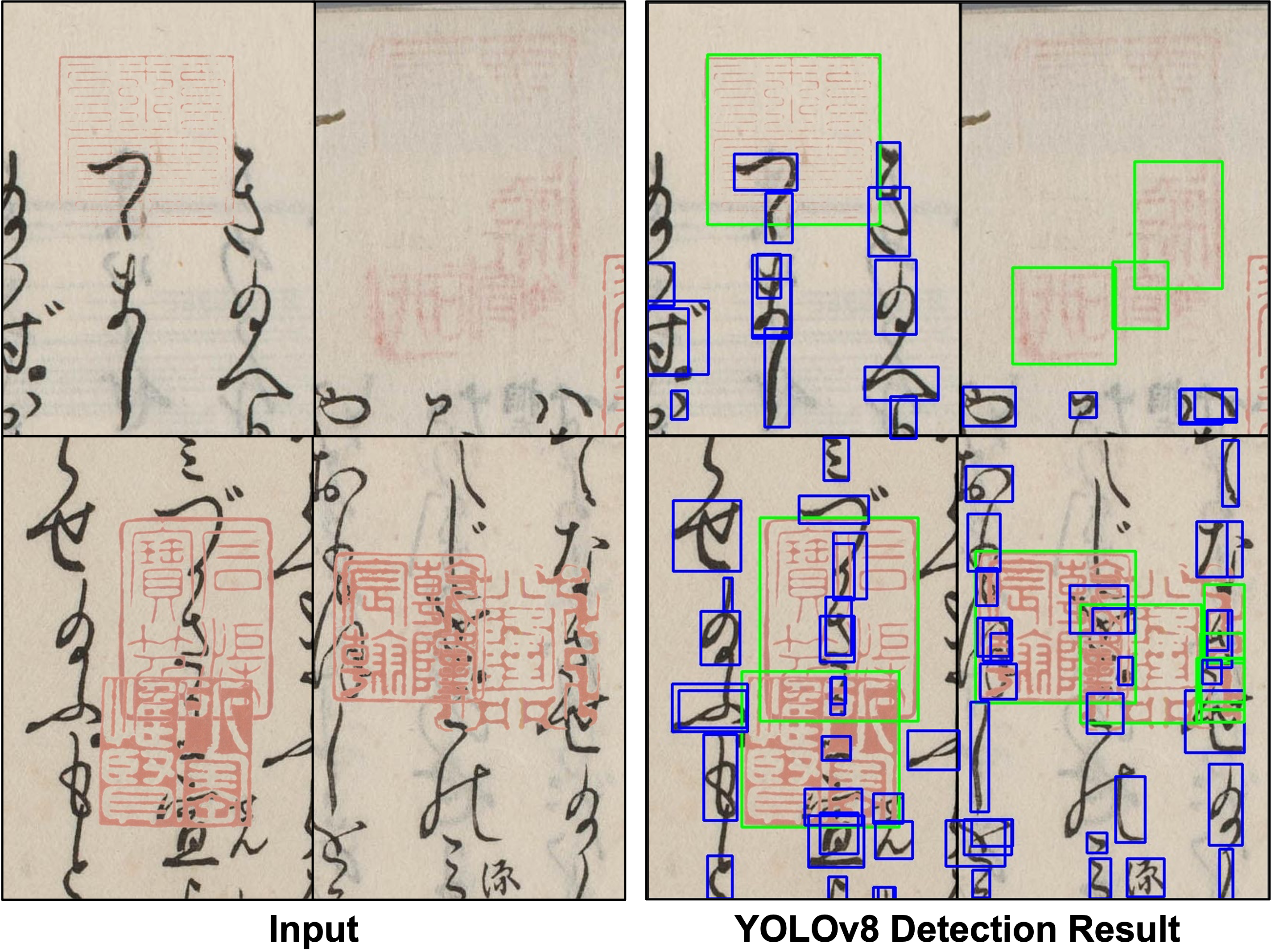}
\vspace{-8pt}
\caption{Examples of challenging cases in Kuzushiji character and seal detection.
The first row shows seals affected by ink fading, while the second row shows seals overlapping with Kuzushiji characters or other seals.
The examples on the right, produced by an existing model, either miss the seal entirely or detect only a portion of it.}
\label{fig_sealoverlap}
\vspace{-2pt}
\end{figure}

To demonstrate the realism of the synthetic Kuzushiji document images, we present a real Kuzushiji document with seals alongside the corresponding synthetic document, as shown in Fig.~\ref{fig_synthetic}.
In the left examples, all seals are real, whereas the additional seals in the right examples are added synthetically.
The comparison shows that the synthetic images appear visually realistic and exhibit seal placement and appearance consistent with real documents.

We constructed the easy-level (Testing-E) and difficult-level (Testing-D) subsets of the test set according to the number of seals per document. 
As shown in Table~\ref{tab:statistics}, each document image in Testing-E contains an average of 2.8 seals, while each document image in Testing-D contains an average of 14.0 seals.

To generate the binarization ground-truth, we employed a combined approach of neural network-based binarization followed by manual verification and correction.
As shown in Fig.~\ref{fig_workflow}, we first applied a pre-trained binarization model trained on the DIBCO benchmarks to binarize the raw Kuzushiji document images (without added seals), producing the initial binarization ground-truth.

Subsequently, a trained Kuzushiji expert manually verified these initial results. 
Even in the absence of seal interference, the initial binarization often contained errors. 
As shown in Fig.~\ref{fig_verification}, the expert identified issues such as missing marginal notes, voiced sound symbols, periods, and unremoved stains.
These errors were then manually corrected and re-verified by another researcher to produce the final binarization ground-truth.
This verification step was essential and required a trained Kuzushiji expert to ensure the accuracy of the ground-truth annotations.

\subsection{Detection Annotations}
The detection annotations include bounding boxes for both Kuzushiji characters and seals.
The bounding box information for Kuzushiji characters was obtained from the OCR annotations provided by CODH~\cite{genjimonogatari}, while the bounding boxes for the seals were recorded during the process of randomly adding them to the Kuzushiji document images.
Notably, the OCR annotations from CODH~\cite{genjimonogatari} include only the main Kuzushiji text, excluding marginal notes.
Considering practical applications, we followed the same strategy.

All annotations are provided in the YOLO format, enabling researchers to directly download and use the dataset.
In addition, detailed information for both Kuzushiji characters and seals, including their coordinates and sizes, is available on our GitHub repository, allowing researchers to convert the data into other object detection formats for model training and evaluation.

\begin{table*}[t]
\centering
\caption{Quantitative comparison of medium-sized YOLO models for Kuzushiji character and seal detection on the DKDS Testing-E subset.
``P'', ``R'', and ``F'' indicate precision, recall, and F-measure (all in \%), respectively.}
\setlength{\tabcolsep}{8pt}{
\begin{tabular}{lcccccccc}
\toprule
\multirow{2}{*}{\textbf{Model}} & \multirow{2}{*}{\textbf{Params}} & \multirow{2}{*}{\textbf{FLOPs}} & \multicolumn{3}{c}{\textbf{Kuzushiji}} & \multicolumn{3}{c}{\textbf{Seal}} \\ \cmidrule{4-6} \cmidrule{7-9}
 &  &  & \textbf{P} & \textbf{R} & \textbf{F} & \textbf{P} & \textbf{R} & \textbf{F} \\ \midrule
YOLOv8 (Ultralytics)~\cite{jocher2023yolov8} & 25.84M & 78.7G & 96.0 & 89.2 & 92.5 & 95.5 & 96.4 & 95.9 \\
YOLOv9 (ECCV2024)~\cite{wang2024yolov9} & 20.01M & 76.5G & 96.6 & 89.6 & 93.0 & 99.8 & 96.4 & 98.1 \\
YOLOv10 (NeurIPS2024)~\cite{wang2024yolov10} & 15.31M & 58.9G & 94.1 & 89.6 & 91.8 & 99.6 & 96.4 & 98.0 \\
YOLO11 (Ultralytics)~\cite{jocher2024yolo11} & 20.03M & 67.7G & 97.7 & 92.0 & 94.8 & 97.0 & 92.9 & 94.9 \\ \bottomrule
\end{tabular}}
\label{tab:detection_easy}
\vspace{-2pt}
\end{table*}

\section{Track 1}\label{sec:track_1}
\subsection{Task Definition}
We define the first track of the proposed dataset as Kuzushiji character and seal detection.
As shown in Fig.~\ref{fig_sealoverlap}, seals may overlap with Kuzushiji characters or even with other seals, and may also be affected by ink fading.
These degradations can lead to incomplete seal detection and significantly reduce the accuracy of Kuzushiji character detection, making the task particularly challenging.

Furthermore, this task has important research significance.
Accurate detection of Kuzushiji characters and seals serves as a crucial preliminary step for downstream applications, such as Kuzushiji character recognition and seal analysis.

\subsection{Evaluation Metrics}
For the Kuzushiji character and seal detection task (Track 1), we adopt standard evaluation metrics commonly used in text detection~\cite{karatzas2015icdar,liao2020real}, including the number of model parameters (Params), floating-point operations (FLOPs), precision (P), recall (R), and F-measure (F).
The number of model parameters reflects the model size and complexity, while FLOPs quantify the computational cost.
Precision (P), recall (R), and F-measure (F) provide a comprehensive evaluation of Kuzushiji character and seal detection performance.

\subsection{Baselines}
For the Kuzushiji character and seal detection task (Track 1), we employ the YOLO series of object detection models, including YOLOv8~\cite{jocher2023yolov8}, YOLOv9~\cite{wang2024yolov9}, YOLOv10~\cite{wang2024yolov10}, and YOLO11~\cite{jocher2024yolo11}.
As one-stage detectors, these models maintain relatively low computational complexity and model size while achieving high detection accuracy, providing a good balance between inference speed and performance. 
In addition, they support end-to-end training, provide fast inference, and exhibit strong generalization capability, making them widely adopted in object detection tasks.

\begin{table*}[t]
\centering
\caption{Quantitative comparison of medium-sized YOLO models for Kuzushiji character and seal detection on the DKDS Testing-D subset.
``P'', ``R'', and ``F'' indicate precision, recall, and F-measure (all in \%), respectively.}
\setlength{\tabcolsep}{8pt}{
\begin{tabular}{lcccccccc}
\toprule
\multirow{2}{*}{\textbf{Model}} & \multirow{2}{*}{\textbf{Params}} & \multirow{2}{*}{\textbf{FLOPs}} & \multicolumn{3}{c}{\textbf{Kuzushiji}} & \multicolumn{3}{c}{\textbf{Seal}} \\ \cmidrule{4-6} \cmidrule{7-9}
 &  &  & \textbf{P} & \textbf{R} & \textbf{F} & \textbf{P} & \textbf{R} & \textbf{F} \\ \midrule
YOLOv8 (Ultralytics)~\cite{jocher2023yolov8} & 25.84M & 78.7G & 91.2 & 87.7 & 89.4 & 92.9 & 81.4 & 86.8 \\
YOLOv9 (ECCV2024)~\cite{wang2024yolov9} & 20.01M & 76.5G & 94.7 & 86.8 & 90.6 & 88.4 & 81.5 & 84.8 \\
YOLOv10 (NeurIPS2024)~\cite{wang2024yolov10} & 15.31M & 58.9G & 90.9 & 86.7 & 88.8 & 96.4 & 77.1 & 85.7 \\
YOLO11 (Ultralytics)~\cite{jocher2024yolo11} & 20.03M & 67.7G & 95.3 & 88.7 & 91.9 & 98.4 & 84.3 & 90.8 \\ \bottomrule
\end{tabular}}
\label{tab:detection_difficult}
\vspace{-2pt}
\end{table*}

\begin{table*}[t]
\centering
\caption{Quantitative comparison of medium-sized YOLO models for Kuzushiji character and seal detection on the DKDS Testing-R subset.
``P'', ``R'', and ``F'' indicate precision, recall, and F-measure (all in \%), respectively.}
\setlength{\tabcolsep}{8pt}{
\begin{tabular}{lcccccccc}
\toprule
\multirow{2}{*}{\textbf{Model}} & \multirow{2}{*}{\textbf{Params}} & \multirow{2}{*}{\textbf{FLOPs}} & \multicolumn{3}{c}{\textbf{Kuzushiji}} & \multicolumn{3}{c}{\textbf{Seal}} \\ \cmidrule{4-6} \cmidrule{7-9}
 &  &  & \textbf{P} & \textbf{R} & \textbf{F} & \textbf{P} & \textbf{R} & \textbf{F} \\ \midrule
YOLOv8 (Ultralytics)~\cite{jocher2023yolov8} & 25.84M & 78.7G & 84.1 & 83.3 & 83.7 & 75.8 & 85.7 & 80.4 \\
YOLOv9 (ECCV2024)~\cite{wang2024yolov9} & 20.01M & 76.5G & 91.6 & 82.1 & 86.6 & 75.5 & 77.3 & 76.4 \\
YOLOv10 (NeurIPS2024)~\cite{wang2024yolov10} & 15.31M & 58.9G & 86.3 & 79.5 & 82.8 & 65.2 & 77.3 & 70.7 \\
YOLO11 (Ultralytics)~\cite{jocher2024yolo11} & 20.03M & 67.7G & 85.4 & 85.3 & 85.3 & 72.3 & 83.0 & 77.3 \\ \bottomrule
\end{tabular}}
\label{tab:detection_real}
\vspace{-2pt}
\end{table*}

\begin{figure*}[t]
\centering
\includegraphics[width=\linewidth]{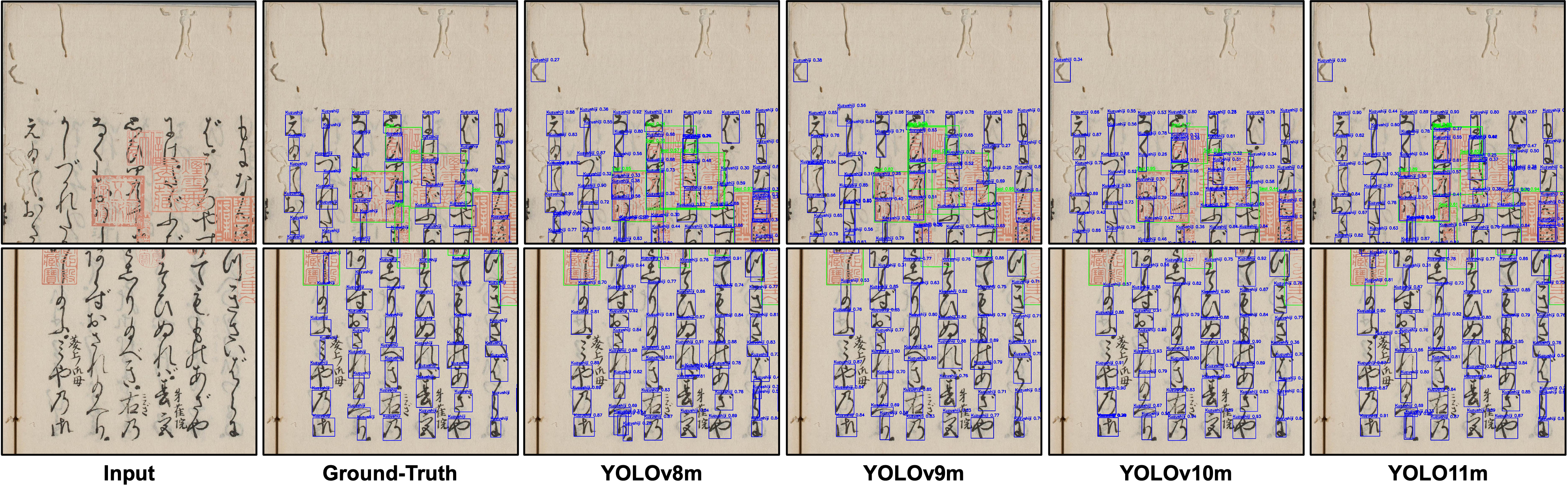}
\vspace{-8pt}
\caption{Visual comparison of Kuzushiji character and seal detection results predicted by different models, together with the corresponding ground-truth annotations. 
The input image is ``200003803\_00028\_2'' from the DKDS Testing-D subset.}
\label{fig_detection1}
\vspace{-2pt}
\end{figure*}

\begin{figure*}[t]
\centering
\includegraphics[width=\linewidth]{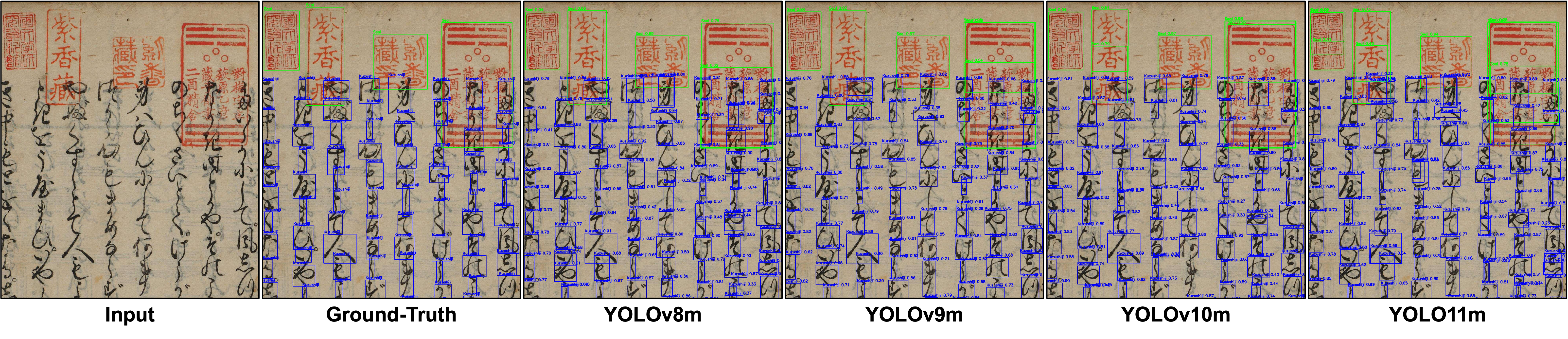}
\vspace{-8pt}
\caption{Visual comparison of Kuzushiji character and seal detection results predicted by different models, together with the corresponding ground-truth annotations. 
The input image is ``200020019\_00002\_2'' from the DKDS Testing-R subset.}
\label{fig_detection2}
\vspace{-2pt}
\end{figure*}

\section{Experiments on Track 1}\label{sec:experiment_1}
\subsection{Implementation Details}
To ensure a fair performance comparison for track 1 on the DKDS dataset, all models were trained and evaluated on the same hardware (i.e., NVIDIA GeForce RTX 3090 GPU). 
All experiments were conducted using the Ubuntu operating system, Python 3.9, and the PyTorch framework.

For the Kuzushiji character and seal detection task (Track 1), the YOLO models were trained for 100 epochs with a batch size of 64 and an input resolution of $640\times640$.
Model training was performed using the SGD optimizer~\cite{robbins1951stochastic} with an initial learning rate of 0.01.
All YOLO models were initialized with weights pre-trained on the COCO dataset~\cite{lin2014microsoft}.

\subsection{Quantitative Comparison}
We trained and evaluated the YOLO series models on the Testing-E, Testing-D, and Testing-R subsets of the DKDS dataset, using an input size of $640\times640$ and the medium-sized variant for all models.
The quantitative comparison results are reported in Tables~\ref{tab:detection_easy}, \ref{tab:detection_difficult}, and~\ref{tab:detection_real}, respectively.

Regarding model efficiency, YOLOv10 had the smallest parameter count at 15.31M and the lowest computational cost at 58.9G FLOPs.
Meanwhile, YOLO11 demonstrated an excellent balance between efficiency (i.e., 20.03M Params and 67.7G FLOPs) and performance while maintaining high detection accuracy.

\subsubsection{Testing-E}
On the Testing-E subset, all models demonstrate strong detection performance, as shown in Table~\ref{tab:detection_easy}.
For Kuzushiji character detection, YOLO11 achieves the best results with an F-measure of 94.8, balancing a precision of 97.7 and a recall of 92.0, and significantly outperforming YOLOv8, YOLOv9, and YOLOv10.
In contrast, for seal detection, YOLOv9 achieves the highest F-measure of 98.1, closely followed by YOLOv10 at 98.0 and YOLOv8 at 95.9, while YOLO11 shows comparatively lower performance with an F-measure of 94.9.

\subsubsection{Testing-D}
The Testing-D subset contains a larger number of seals per document, which markedly increases the overall detection difficulty, as shown in Table~\ref{tab:detection_difficult}.
For Kuzushiji character detection, YOLO11 continues to achieve the highest F-measure of 91.9, demonstrating robust character modeling and detection capability even under complex degradation conditions.
For seal detection, YOLO11 shows an even more pronounced advantage, attaining an F-measure of 90.8 and clearly outperforming the other models, including YOLOv8 at 86.8, YOLOv9 at 84.8, and YOLOv10 at 85.7.
Compared with Testing-E, the recall of seal detection decreases markedly for all models on Testing-D, indicating that incomplete seal detection is the primary source of the increased difficulty observed in Testing-D.

\subsubsection{Testing-R}
For the Testing-R subset, although the number of seals per document image is relatively low, the detection task remains highly challenging due to variations in document layouts, seal shapes, and Kuzushiji writing styles, as shown in Table~\ref{tab:detection_real}.
Specifically, all models perform worse than on the Testing-E and Testing-D subsets.
In Kuzushiji character detection, YOLOv9 achieves the best performance with a precision of 91.6, a recall of 82.1, and an F-measure of 86.6.
For seal detection, YOLOv8 demonstrates the best performance with a precision of 75.8, a recall of 85.7, and an F-measure of 80.4.

\subsection{Visualization}
We present the detection results of various YOLO models on the Testing-D subset for Kuzushiji character and seal detection, together with the corresponding input images and ground-truth annotations.
As shown in Fig.~\ref{fig_detection1}, seals are marked with green bounding boxes, while Kuzushiji characters are marked with blue bounding boxes.
As shown in the first row, all YOLO models incorrectly detect stains in the upper-left corner as Kuzushiji characters, which constitutes a primary factor affecting detection performance.
In the second row, YOLOv8m repeatedly detects the same character in the bottom-left corner.
Furthermore, although the ground-truth annotations do not include positional information for marginal notes, YOLO11m still detects several marginal notes to varying degrees.

Fig.~\ref{fig_detection2} presents the detection results of different models on the Testing-R subset.
We can observe that while all models can detect seals and Kuzushiji characters, they show varying degrees of duplicate seal detection. 
This is one of the reasons for the lower quantitative results of these models.

\section{Track 2}\label{sec:track_2}
\subsection{Task Definition}
\begin{figure}[t]
\centering
\includegraphics[width=\linewidth]{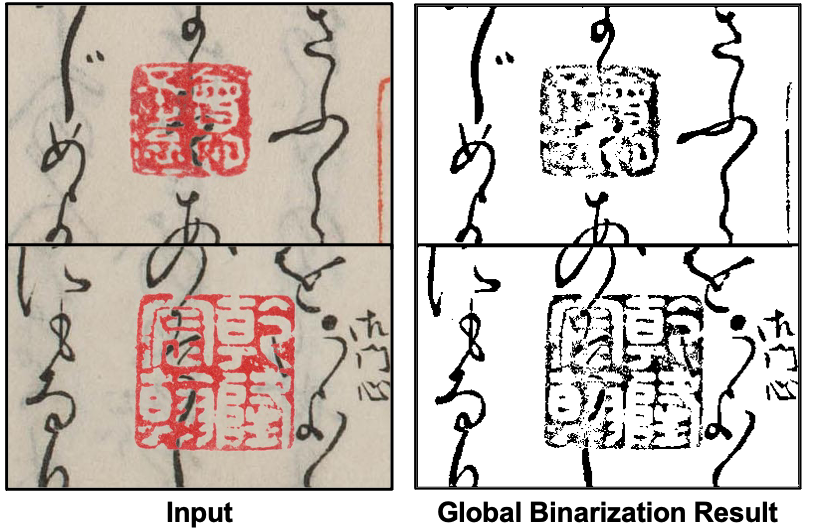}
\vspace{-8pt}
\caption{Challenging examples of document binarization involving overlaps between Kuzushiji characters and seals. 
The example output on the right, generated by an existing algorithm, retains noise from the seal, resulting in unsatisfactory performance.}
\label{fig_kuzushijioverlap}
\vspace{-2pt}
\end{figure}

We define the second track of the proposed dataset as document binarization.
In this track, the objective is to remove seals while preserving, or even restoring, the underlying Kuzushiji characters.
This task becomes particularly challenging when characters and seals overlap, as shown in Fig.~\ref{fig_kuzushijioverlap}.

In addition, this task plays an important role in document image analysis.
Accurate document binarization can improve the accuracy of subsequent Kuzushiji character recognition systems.

\subsection{Evaluation Metrics}
For the document binarization task (Track 2), we employ standard evaluation metrics~\cite{gatos2009icdar} for quantitative comparison, including the F-measure (FM), pseudo F-measure (p-FM), peak signal-to-noise ratio (PSNR), and distance reciprocal distortion (DRD), which measure pixel-level classification accuracy, text-region preservation capability, overall reconstruction quality, and perceptual distortion, respectively.

To provide a more comprehensive assessment, we also adopt the Average-Score Metric (ASM) proposed by Ju \emph{et al.}~\cite{ju2025efficient,ju2025mfegan}, which is calculated as follows:
\begin{equation}
\text{ASM}\!=\!\frac{1}{4} (\text{FM}\!+\!\text{p-FM}\!+\!\text{PSNR}\!+\!(100\!-\!\text{DRD)}),
\end{equation}
where a higher ASM value indicates better overall performance and provides a single unified metric.

\subsection{Baselines}
For the document binarization task (Track 2), we employ several traditional binarization algorithms on Kuzushiji document images, including Niblack~\cite{niblack1985introduction}, Otsu~\cite{otsu1979threshold}, and Sauvola~\cite{sauvola2000adaptive}, which can be directly applied without any training.

Considering that seals in pre-modern Japanese documents are typically red, while Kuzushiji characters are black and the paper background is pale yellow, we introduce a preprocessing step to remove seal regions based on K-means clustering~\cite{mcqueen1967some} before applying the binarization algorithms.
This preprocessing effectively reduces the interference caused by red seals and enhances the quality of subsequent binarization results.
Specifically, we implement three combined approaches: K-means clustering + Otsu, K-means clustering + Niblack, and K-means clustering + Sauvola.

To provide deep learning-based baselines, we employ two SOTA GAN-based methods (i.e., Suh \emph{et al.}~\cite{suh2022two} and Ju \emph{et al.}~\cite{ju2024three}) to evaluate Track 2 of the proposed DKDS dataset.

\begin{figure}[t]
\centering
\includegraphics[width=\linewidth]{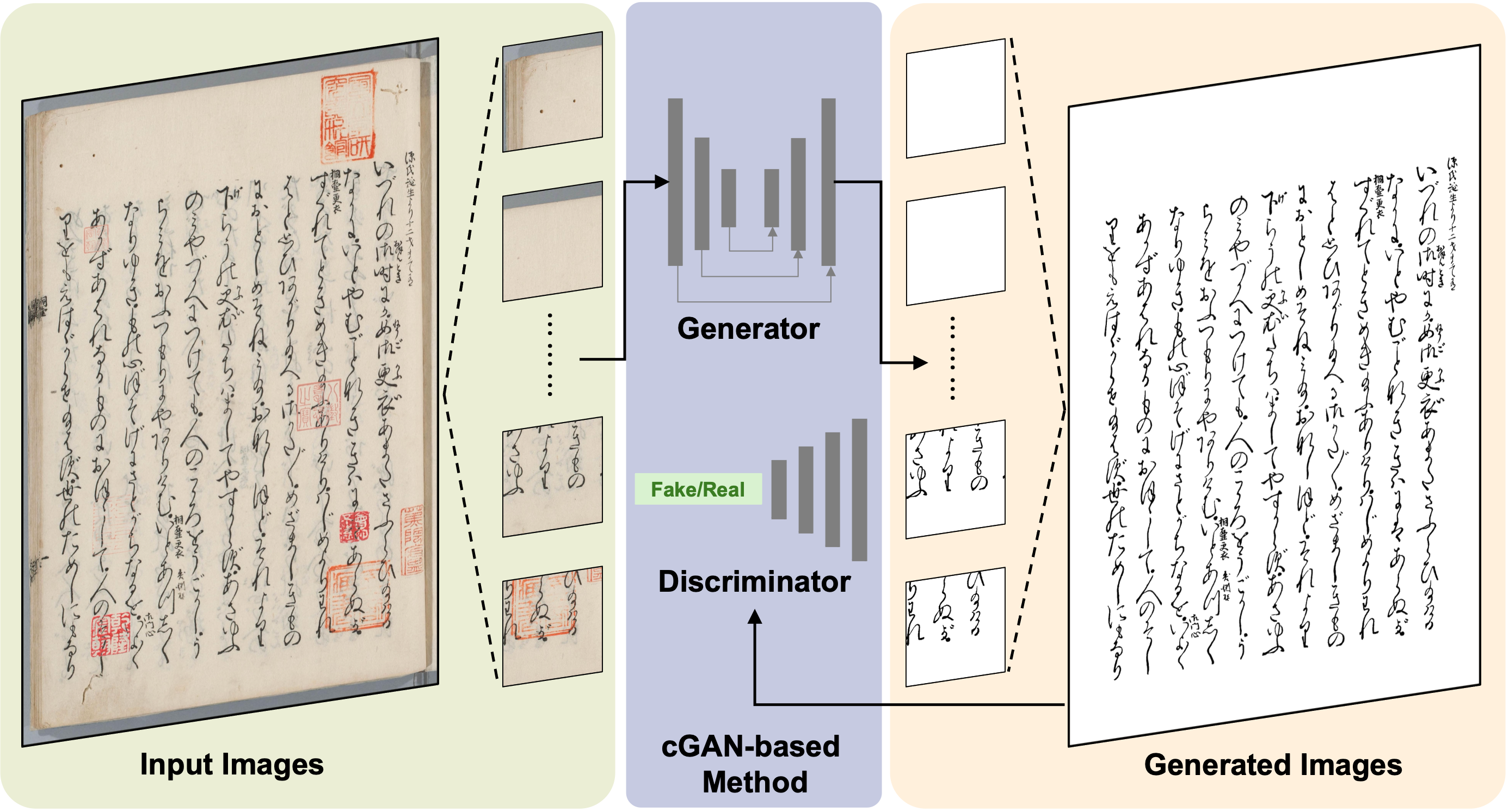}
\vspace{-8pt}
\caption{The architecture diagram of our improved conditional GAN (cGAN)-based method.}
\label{fig_gan}
\vspace{-2pt}
\end{figure}

In addition, we employ an improved cGAN~\cite{isola2017image}-based method to generate binarized Kuzushiji document images.
The network architecture for this model is shown in Fig.~\ref{fig_gan}.
Since directly inputting the entire image (approximately $2100\times3200$ pixels) would require excessive GPU memory, we divide each document image into multiple $512\times512$ patches; the details of this division are described in Section~\ref{sec:details_binarization}.
These patches are then fed into a U-Net~\cite{ronneberger2015u} architecture with an EfficientNet-B5~\cite{tan2019rethinking} backbone, which serves as the generator.
A simplified PatchGAN~\cite{isola2017image} discriminator is adopted to distinguish between real and fake (generated) images.
Furthermore, following Suh \emph{et al.}~\cite{suh2022two} and Ju \emph{et al.}~\cite{ju2024three}, we employ a loss function that combines the Wasserstein GAN with Gradient Penalty (WGAN-GP) loss~\cite{gulrajani2017improved} and an additional Binary Cross-Entropy (BCE) loss.

\begin{table*}[t]
\centering
\caption{Quantitative comparison of various methods for document binarization on the DKDS Testing-E subset.}
\setlength{\tabcolsep}{9.5pt}{
\begin{tabular}{lcccccc}
\toprule
\textbf{Method} & \textbf{Params} & \textbf{FM$\uparrow$} & \textbf{p-FM$\uparrow$} & \textbf{PSNR$\uparrow$} & \textbf{DRD$\downarrow$} & \textbf{ASM$\uparrow$} \\ \midrule
Otsu~\cite{otsu1979threshold} & -- & 63.01 & 63.31 & 11.76dB & 37.69 & 50.10 \\
Niblack~\cite{niblack1985introduction}  & -- & 39.13 & 41.14 & 8.44dB & 79.70 & 27.25 \\
Sauvola~\cite{sauvola2000adaptive}  & -- & 87.87 & 90.99 & 18.34dB & 7.01 & 72.55 \\ \midrule
K-means~\cite{mcqueen1967some} + Otsu~\cite{otsu1979threshold}  & -- & 84.76 & 86.28 & 17.14dB & 9.90 & 69.57 \\
K-means~\cite{mcqueen1967some} + Niblack~\cite{niblack1985introduction}  & -- & 39.99 & 42.03 & 8.61dB & 76.67 & 28.49 \\
K-means~\cite{mcqueen1967some} + Sauvola~\cite{sauvola2000adaptive}  & -- & 88.59 & 91.48 & 18.65dB & 6.37 & 73.09 \\ \midrule
Suh \emph{et al.} (PR2022)~\cite{suh2022two} & 187.30M & 93.09 & 93.09 & 20.99dB & 3.12 & 76.01 \\ 
Ju \emph{et al.} (KBS2024)~\cite{ju2024three} & 191.46M & 95.80 & 95.81 & 23.14dB & 1.76 & 78.25 \\
Ours & 31.22M & 98.11 & 98.14 & 26.53dB & 0.82 & 80.49 \\ \bottomrule
\end{tabular}}
\label{tab:binarization_easy}
\vspace{-2pt}
\end{table*}

\begin{table*}[t]
\centering
\caption{Quantitative comparison of various methods for document binarization on the DKDS Testing-D subset.}
\setlength{\tabcolsep}{9.5pt}{
\begin{tabular}{lcccccc}
\toprule
\textbf{Method} & \textbf{Params} & \textbf{FM$\uparrow$} & \textbf{p-FM$\uparrow$} & \textbf{PSNR$\uparrow$} & \textbf{DRD$\downarrow$} & \textbf{ASM$\uparrow$} \\ \midrule
Otsu~\cite{otsu1979threshold} & -- & 60.51 & 60.80 & 11.39dB & 40.89 & 47.95 \\
Niblack~\cite{niblack1985introduction}  & -- & 37.52 & 39.41 & 8.28dB & 82.29 & 25.73 \\
Sauvola~\cite{sauvola2000adaptive}  & -- & 84.00 & 87.04 & 17.08dB & 9.60 & 69.63 \\ \midrule
K-means~\cite{mcqueen1967some} + Otsu~\cite{otsu1979threshold}  & -- & 68.10 & 69.04 & 13.37dB & 33.32 & 54.30 \\
K-means~\cite{mcqueen1967some} + Niblack~\cite{niblack1985introduction}  & -- & 35.98 & 37.04 & 8.27dB & 82.66 & 24.66 \\
K-means~\cite{mcqueen1967some} + Sauvola~\cite{sauvola2000adaptive}  & -- & 70.50 & 69.98 & 15.80dB & 14.69 & 60.40 \\ \midrule
Suh \emph{et al.} (PR2022)~\cite{suh2022two} & 187.30M & 91.95 & 91.96 & 20.10dB & 3.85 & 75.04 \\ 
Ju \emph{et al.} (KBS2024)~\cite{ju2024three} & 191.46M & 94.86 & 94.90 & 22.11dB & 2.27 & 77.40 \\
Ours & 31.22M & 97.08 & 97.13 & 24.58dB & 1.38 & 79.35 \\ \bottomrule
\end{tabular}}
\label{tab:binarization_difficult}
\vspace{-2pt}
\end{table*}

\begin{table*}[t]
\centering
\caption{Quantitative comparison of various methods for document binarization on the DKDS Testing-R subset.}
\setlength{\tabcolsep}{9.5pt}{
\begin{tabular}{lcccccc}
\toprule
\textbf{Method} & \textbf{Params} & \textbf{FM$\uparrow$} & \textbf{p-FM$\uparrow$} & \textbf{PSNR$\uparrow$} & \textbf{DRD$\downarrow$} & \textbf{ASM$\uparrow$} \\ \midrule
Otsu~\cite{otsu1979threshold} & -- & 67.85 & 69.36 & 13.05dB & 52.82 & 49.36 \\
Niblack~\cite{niblack1985introduction}  & -- & 39.67 & 42.01 & 8.31dB & 81.81 & 27.04 \\
Sauvola~\cite{sauvola2000adaptive}  & -- & 73.69 & 76.14 & 14.30dB & 32.84 & 57.82 \\ \midrule
K-means~\cite{mcqueen1967some} + Otsu~\cite{otsu1979threshold}  & -- & 48.92 & 49.39 & 8.60dB & 77.09 & 32.46 \\
K-means~\cite{mcqueen1967some} + Niblack~\cite{niblack1985introduction}  & -- & 33.68 & 34.43 & 8.16dB & 83.82 & 23.11 \\
K-means~\cite{mcqueen1967some} + Sauvola~\cite{sauvola2000adaptive}  & -- & 53.00 & 52.80 & 12.98dB & 25.54 & 48.31 \\ \midrule
Suh \emph{et al.} (PR2022)~\cite{suh2022two} & 187.30M & 81.00 & 81.53 & 16.26dB & 16.10 & 65.67 \\ 
Ju \emph{et al.} (KBS2024)~\cite{ju2024three} & 191.46M & 77.37 & 77.55 & 15.47dB & 30.55 & 59.96 \\
Ours & 31.22M & 73.33 & 72.98 & 15.74dB & 31.44 & 57.65 \\ \bottomrule
\end{tabular}}
\label{tab:binarization_real}
\vspace{-2pt}
\end{table*}

\section{Experiments on Track 2}\label{sec:experiment_2}
\subsection{Implementation Details}\label{sec:details_binarization}
All experiments were carried out on an Ubuntu system using Python 3.9 and the PyTorch framework.
For Track 2 of the dataset, all models were trained and evaluated under the same hardware configuration (i.e., NVIDIA GeForce RTX 3090 GPU) to ensure a fair performance comparison.

For the document binarization task (Track 2), the input images were first converted into grayscale as a preprocessing step for traditional binarization algorithms (i.e., Niblack~\cite{niblack1985introduction}, Otsu~\cite{otsu1979threshold}, and Sauvola~\cite{sauvola2000adaptive}).
The Niblack algorithm used a $25\times25$ sliding window with $k=0.8$, while the Sauvola algorithm employed the same window size.

For the K-means clustering~\cite{mcqueen1967some}, the number of clusters was set to $K=3$.
After clustering, median filtering with a kernel size of $5\times5$ was applied to smooth the results and remove isolated noise points.

For the two SOTA GAN-based methods, we followed their original training data preprocessing strategies, resulting in 325,608 patches of size $256\times256$ pixels and 240 downscaled full-page images of $512\times512$ pixels.
In terms of model architecture, we adopted a U-Net~\cite{ronneberger2015u} generator for the method of Suh \emph{et al.}~\cite{suh2022two} and employed a U-Net++~\cite{zhou2019unet++} generator for the method of Ju \emph{et al.}~\cite{ju2024three}.
Both methods used EfficientNet-B5~\cite{tan2019rethinking} as the encoder backbone.

For our cGAN~\cite{isola2017image}-based method, the training data were preprocessed by dividing each original image into $512\times512$ overlapping patches with a 30\% overlap ratio.
To further expand the training set, we applied data augmentation, including scaling (factors 0.75, 1.0, 1.25, and 1.5) and rotations (0\textdegree and 270\textdegree).
After preprocessing and augmentation, the 40 original training images were expanded into a total of 71,458 patches of size $512\times512$ pixels.
Both the generator and discriminator were optimized using the Adam~\cite{kingma2014adam} optimizer with a learning rate of $2\times10^{-4}$ and $\beta$ coefficients of $(0.5, 0.999)$.
The model was trained for 10 epochs with a batch size of 16.
During inference, the predicted outputs were binarized using a threshold of 0.5.

\subsection{Quantitative Comparison}
We employed traditional binarization algorithms, traditional algorithms combined with K-means clustering, two SOTA GAN-based methods, and our improved cGAN~\cite{isola2017image}-based method as baselines for the document binarization task.
The experimental results are summarized in Tables~\ref{tab:binarization_easy}, \ref{tab:binarization_difficult}, and~\ref{tab:binarization_real}, respectively.

\subsubsection{Testing-E}
On the Testing-E subset, the traditional binarization algorithms exhibit generally poor performance, as shown in Table~\ref{tab:binarization_easy}.
Specifically, Niblack's method~\cite{niblack1985introduction} performs the worst, as indicated by its high DRD value of 79.70, reflecting strong sensitivity to noise in the dataset.
Otsu's method~\cite{otsu1979threshold}, a simple global thresholding method, achieves moderate results (i.e., an ASM value of 50.10), while Sauvola's method~\cite{sauvola2000adaptive} significantly outperforms both by leveraging locally adaptive thresholding to process complex backgrounds more effectively, achieving an ASM value of 72.55.

Furthermore, applying K-means clustering~\cite{mcqueen1967some} as a preprocessing step to remove red-colored interference (e.g., seals) further improves performance.
As shown in Table~\ref{tab:binarization_easy}, the ASM increases from 27.25 to 28.49, from 50.10 to 69.57, and from 72.55 to 73.09 for K-means clustering + Niblack, K-means clustering + Otsu, and K-means clustering + Sauvola, respectively.
These results indicate that when the number of seals per document is limited, color-based clustering effectively mitigates seal interference for traditional thresholding methods.

However, when the traditional methods are applied to the Testing-D subset, which contains a larger number of seals, the limitations of combining K-means clustering with thresholding methods become more evident.
Although K-means preprocessing still provides some improvement for simple thresholding approaches, overall performance deteriorates significantly.
In particular, the FM, p-FM, and DRD metrics for both K-means clustering + Otsu and K-means clustering + Sauvola exhibit noticeable degradation.
This indicates that in more complex scenarios, K-means clustering-based binarization strategies struggle to robustly process the effects of severe degradation.

\begin{figure*}[t]
\centering
\includegraphics[width=\linewidth]{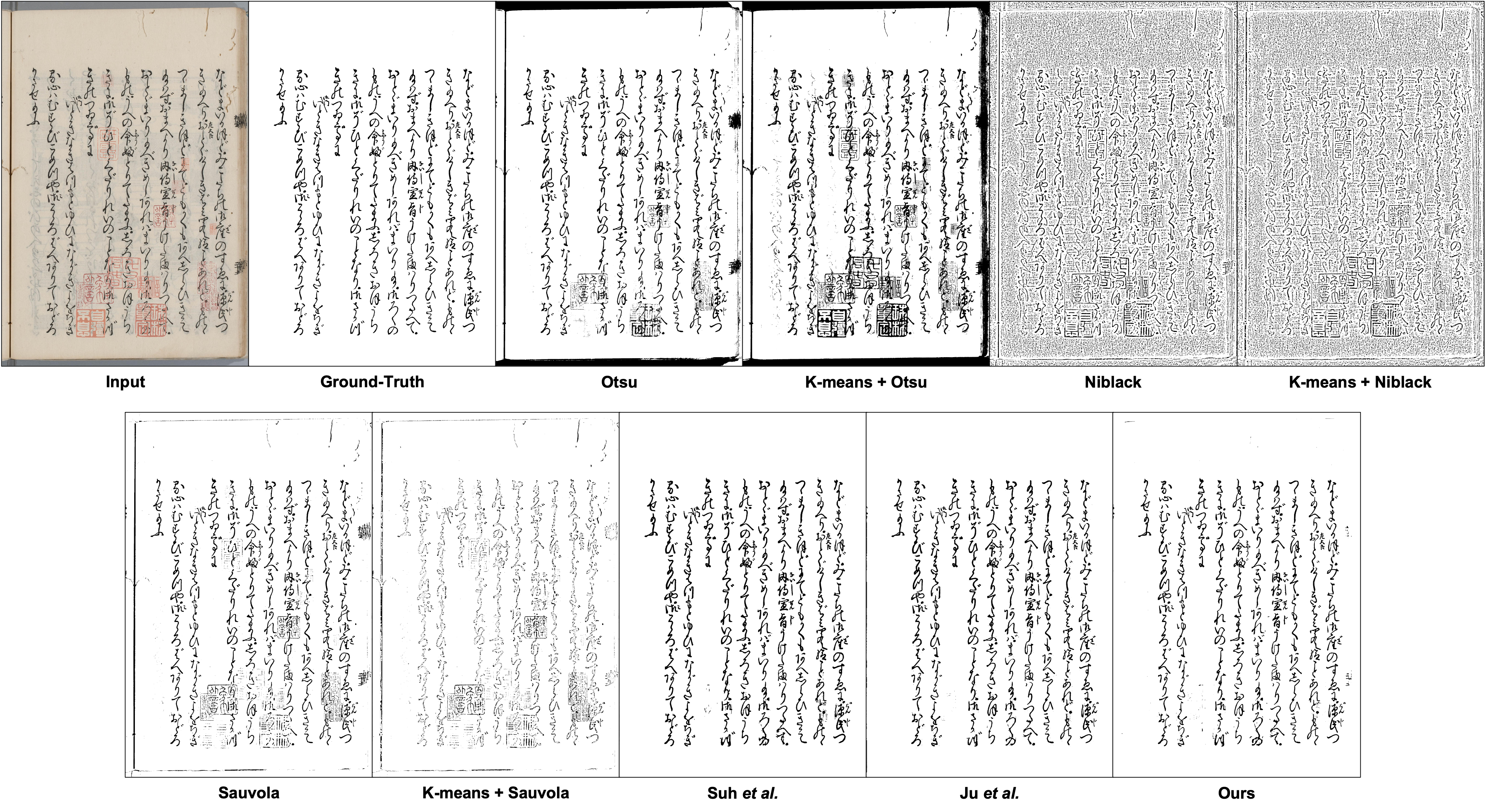}
\vspace{-8pt}
\caption{Visual comparison of document binarization results produced by different methods on the Kuzushiji document ``200003803\_00027\_1'' from the DKDS Testing-D subset. 
The input image, ground-truth image, and the binarization results obtained by different methods are shown.}
\label{fig_visual1}
\vspace{-2pt}
\end{figure*}

\begin{figure*}[t]
\centering
\includegraphics[width=\linewidth]{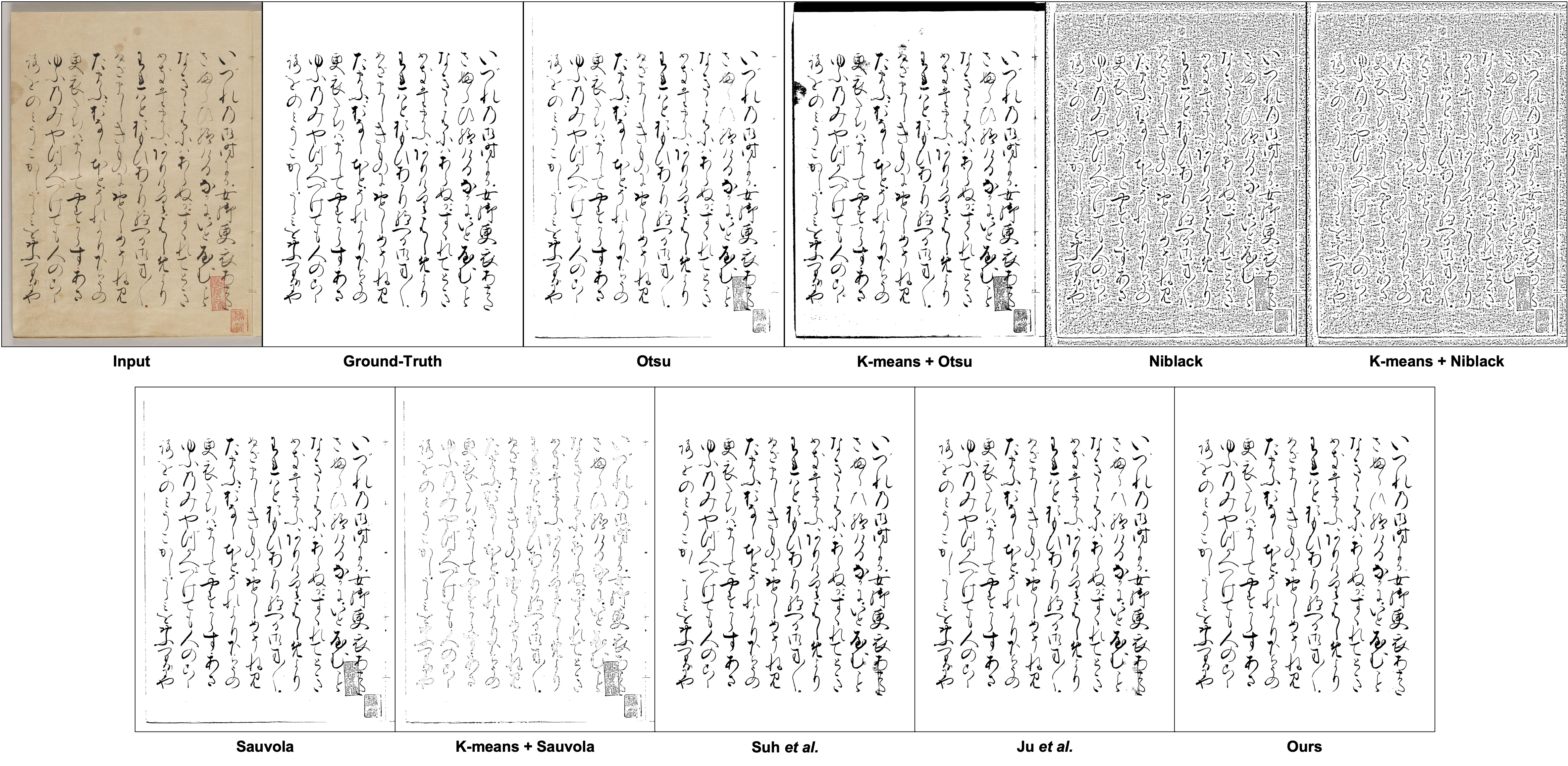}
\vspace{-8pt}
\caption{Visual comparison of document binarization results produced by different methods on the Kuzushiji document ``200010454\_00002\_2'' from the DKDS Testing-R subset. 
The input image, ground-truth image, and the binarization results obtained by different methods are shown.}
\label{fig_visual2}
\vspace{-2pt}
\end{figure*}

In contrast, deep learning-based methods consistently outperform traditional methods.
On the Testing-E subset, the method of Suh \emph{et al.}~\cite{suh2022two} achieves FM and p-FM values of 93.09 with a DRD of 3.12, while the method of Ju \emph{et al.}~\cite{ju2024three} further improves the performance to FM and p-FM values of 95.80 and 95.81, respectively, and reduces the DRD to 1.76.
Our method achieves the best overall performance across all evaluation metrics, attaining FM and p-FM values of 98.11 and 98.14, a PSNR of 26.53~dB, a DRD of 0.82, and the highest ASM of 80.49.

Finally, in terms of model complexity, the methods of Suh \emph{et al.}~\cite{suh2022two} and Ju \emph{et al.}~\cite{ju2024three} have 187.30M and 191.46M parameters, respectively, whereas our model requires only 31.22M parameters, achieving superior performance with a substantially reduced model size.

\subsubsection{Testing-D}
When evaluated on the more challenging Testing-D subset, the performance of all methods declines.
Specifically, the ASM values of Otsu's method~\cite{otsu1979threshold}, Niblack's method~\cite{niblack1985introduction}, and Sauvola's method~\cite{sauvola2000adaptive} decrease from 50.10 to 47.95, from 27.25 to 25.73, and from 72.55 to 69.63, respectively.
After applying K-means clustering, these three methods still show limited performance.

In contrast, deep learning-based methods still outperform traditional algorithms.
The FM value of the method of Suh \emph{et al.}~\cite{suh2022two} decreases from 93.09 to 91.95, while the method of Ju \emph{et al.}~\cite{ju2024three} decreases from 95.80 to 94.86.

In comparison, our method experiences only a slight decrease from 98.11 to 97.08 while still achieving the lowest DRD value of 1.38. 
Notably, compared with the method of Ju \emph{et al.}~\cite{ju2024three}, our method further reduces the DRD on Testing-D by approximately 39\%, decreasing it from 2.27 to 1.38, while consistently outperforming competing methods in FM, p-FM, and PSNR.

\subsubsection{Testing-R}
For the Testing-R subset, deep learning-based methods still outperform traditional algorithms, as shown in Table~\ref{tab:binarization_real}.
However, our method achieves lower performance than the methods proposed by Suh \emph{et al.}~\cite{suh2022two} and Ju \emph{et al.}~\cite{ju2024three}.
Specifically, the method of Suh \emph{et al.} achieves the highest ASM value of 65.67, while our method achieves an ASM value of 57.65.

We attribute this result to the relatively limited generalization capability of our model, which results in lower performance when processing document images with different layouts, varying Kuzushiji writing styles, and diverse seal shapes.
This result highlights the challenge of generalizing to real-world Kuzushiji documents.

\subsection{Visualization}
We present binarization results produced by different methods on the Testing-D subset, together with the corresponding input and ground-truth images, as shown in Fig.~\ref{fig_visual1}.
The input images contain various forms of degradation, including red seals, paper damage, stains, and ink fading, while the ground-truth images remove all such interference, retaining only the Kuzushiji characters.

Among the traditional methods, Niblack's algorithm~\cite{niblack1985introduction} produces the poorest results, with much noise remaining in the background.
Otsu's method~\cite{otsu1979threshold} achieves better separation between text and background but still retains some noise.
Sauvola's method~\cite{sauvola2000adaptive} performs relatively better but fails to completely remove seal interference.

Furthermore, as shown in Table~\ref{tab:binarization_difficult}, the incorporation of K-means clustering does not lead to a substantial performance improvement when documents contain a large number of seals.

In contrast, deep learning-based methods, including those proposed by Suh \emph{et al.}~\cite{suh2022two}, Ju \emph{et al.}~\cite{ju2024three}, and our method, achieve markedly better results by effectively suppressing noise and seal interference while preserving high text legibility.

In addition, we select a document image from the Testing-R subset to illustrate the binarization results produced by different methods, as shown in Fig.~\ref{fig_visual2}.
In this document image, two seals appear in the lower right corner, one overlapping with Kuzushiji characters while the other does not.

It can be observed that traditional methods (including those combined with K-means clustering) produce unsatisfactory results. 
The methods by Suh \emph{et al.}~\cite{suh2022two} and Ju \emph{et al.}~\cite{ju2024three} fail to completely remove the seals (i.e., some seal remnants remain), while our method effectively removes the seals, although some parts of the overlapping Kuzushiji characters are also removed.

\begin{table}[t]
\centering
\caption{Comparison of OCR performance (character error rate, CER) between degraded and binarized document images (ground-truth) on different subsets.}
\setlength{\tabcolsep}{6pt}
\begin{tabular}{lccc}
\toprule
\textbf{Subset} & \textbf{\begin{tabular}[c]{@{}c@{}}Degraded\\ CER\end{tabular}} & \textbf{\begin{tabular}[c]{@{}c@{}}Binarized\\ CER\end{tabular}} & \textbf{\begin{tabular}[c]{@{}c@{}}Relative \\ Reduction\end{tabular}} \\ \midrule
Training   & 2.3808\% & 1.6681\% & 29.94\% \\
Testing-E  & 0.8128\% & 0.7209\% & 11.31\% \\
Testing-D  & 1.9825\% & 0.7209\% & 63.64\% \\ \bottomrule
\end{tabular}
\label{tab:ocr}
\vspace{-2pt}
\end{table}

\section{OCR Evaluation}\label{sec:ocr_evaluation}
To quantitatively evaluate OCR performance on both degraded and binarized document images, we measured the character error rate (CER, \%) using the ``miwo'' app~\cite{clanuwat2021miwo} for Kuzushiji character recognition.
OCR experiments were conducted on the Training, Testing-E, and Testing-D subsets. 
The results are summarized in Table~\ref{tab:ocr}.

For the Training subset, the CER of the binarized document images was 1.6681\%, representing a 29.94\% reduction compared with the 2.3808\% CER of the corresponding degraded document images.

For the Testing-E and Testing-D subsets, the CERs of the degraded document images were 0.8128\% and 1.9825\%, respectively.
Because the binarized images for these two subsets were obtained from the same ground-truth images, the CER of the binarized images was identical for both subsets (0.7209\%). 
These results indicate that binarization reduced the CER by 11.31\% for Testing-E and 63.64\% for Testing-D.

Notably, both the raw Kuzushiji document images~\cite{genjimonogatari} and the ``miwo'' OCR app~\cite{clanuwat2021miwo} are provided by the Center for Open Data in the Humanities (CODH), which may partly account for the relatively strong baseline recognition performance (i.e., CER on degraded document images) observed in our experiments.

\section{Conclusion}\label{sec:conclusion}
Document binarization of degraded pre-modern Japanese documents presents significant challenges, particularly when seals overlap with Kuzushiji characters.
Effectively removing seal interference while preserving or restoring character details remains a critical problem.
Notably, no publicly available dataset currently contains Kuzushiji documents with seals, limiting the development of methods for this task.

To address this gap, we constructed the Degraded Kuzushiji Documents with Seals (DKDS) dataset by combining pre-modern Kuzushiji document images with high-resolution images of ancient seals.
We detail the construction process of the DKDS dataset and define two task tracks: Kuzushiji character and seal detection, and document binarization.
Furthermore, we provide baseline evaluation results for several classical and deep learning-based methods to facilitate future research.

In addition, the selected Kuzushiji document dataset includes corresponding Unicode character mappings and OCR annotations, providing valuable resources for training models for both document binarization and OCR tasks. 
In future work, we plan to jointly design and train models for these tasks, aiming to develop an end-to-end system capable of efficiently and accurately converting Kuzushiji characters into modern Japanese from degraded pre-modern Japanese documents.

\section*{Acknowledgments}
We sincerely appreciate Jie Chen for kindly providing and supporting the collection of the raw seal data used in this work.

\section*{Funding}
This work was supported by JSPS KAKENHI Grant Number 25H01242, and JST SPRING Grant Number JPMJSP2110.

\section*{Author Contributions}
\textbf{Rui-Yang Ju}: Conceptualization, Data Curation, Formal Analysis, Methodology, Writing - Original Draft Preparation, Writing - Review \& Editing;
\textbf{Kohei Yamashita}: Methodology, Data Curation, Investigation, Writing - Review \& Editing;
\textbf{Hirotaka Kameko}: Project Administration, Resources, Writing - Review \& Editing;
\textbf{Shinsuke Mori}: Funding Acquisition, Resources, Supervision, Writing - Review \& Editing.

\section*{Data availability}
The proposed dataset and implementation code are publicly available on GitHub at~\url{https://github.com/RuiyangJu/DKDS}.

\section*{Declarations}
\subsection*{Competing Interests}
The authors declare that they have no conflict of interest.

\subsection*{Ethics approval}
This research does not involve human participants and/or animals.

\bibliography{reference} 
\end{CJK}
\end{document}